# Language Models Meet Vibration Testing for LEO Constellation Satellites with Calibrated Response Prediction from Few-show Sweeps

**Jun Wu [1,2], Deming Li[1,2], Zhenming Zheng[1,2], Honghao Yue[1,2], Fei Yang[1,2,3], Hao Sun[1,2,†]**

[1] Research Center of Aerospace Mechanism and Control, Harbin Institute of Technology
[2] Laboratory of Aerospace Embodied Intelligence Technology and Equipment, Harbin Institute of Technology
[3] Suzhou Research Institute of HIT
sunnice@hit.edu.cn

[†]Corresponding author.

## Abstract

Large language models (LLMs) increasingly drive engineering analysis end to end, turning textual records into the inputs of numerical programs, but a plausible and wrong value can pass silently into a decision. We propose deterministic validation gates for LLM-driven engineering analysis, a general paradigm in which a typed configuration produced by LLMs is consumed only after it passes deterministic checks on units, physical consistency, verbatim quotes, agreement between independent models, consistency with measurement exports and the applicability of trained models. The gates target hallucinations from two directions, since quote checking catches fabricated content while agreement between models catches errors of omission that quote checking misses. We instantiate the paradigm on sine vibration acceptance testing for low Earth orbit mega-constellation satellites, where DeepSeek V4.1 Flash and Qwen-plus read Chinese test records and extract a configuration of 76 fields. Fact-level F1 ranged from 0.98 to 1.0 on five input variants, with all repairs frozen on development inputs and evaluated on held-out inputs. With all gates, 20 of 20 configurations produced notching results bit for bit identical to the gold run, and all 17 real and injected errors in consumed fields were stopped, of which 16 would have run without gates. Retraining the bound model on a dataset built from an LLM configuration reproduced its results within GPU nondeterminism. We also report a negative result of practical value. Direct LLM curve forecasting only matches linear extrapolation, whereas a small physics-structured network reduces the over-notched bandwidth by 86% to 89% at calibrated coverage c = 0.90, which delineates where direct LLM prediction should give way to a trained model.

## 1. Introduction

Large language models (LLMs) increasingly turn technical text into structured records [1-3] and act as controllers that call tools and programs [4-8]. In engineering analysis the two roles meet. A model reads the conditions of a test from text, and a numerical program uses them to make a decision. Text extraction is usually judged by field-level accuracy, but in such a chain a value that is plausible and wrong matters more than a missing one, because it can pass silently into the decision. We propose deterministic validation gates for LLM-driven engineering analysis as a general paradigm for this setting. A gate is a deterministic check placed between the LLM output and the numerical stage, and a value is consumed only after every gate on its path has passed. The gates address hallucination in both of its common forms, since verbatim quote checking exposes fabricated content while agreement between independent models exposes silent omissions.

We study this question in the acceptance testing of satellites of low Earth orbit (LEO) mega-constellations. Units of one design are produced in series and each passes a vibration campaign. In sine vibration tests the input is reduced, or notched, in frequency bands where a response would exceed its allowable level [9, 10], and the notch profile is usually derived by scaling a low-level sweep to the acceptance input with a fixed margin. The input profile, the response limit, the run sequence, the control scheme and the sensor list are stated in Chinese textual records. In an earlier version of this study these conditions were coded by hand into the analysis. We audited the code and found 20 such conditions, of which 14 can be read from the text.

[*] Code is available at https://gitcode.com/sunhaha1007/llm-vibration-testing

In this study the LLMs drive the analysis. They read the Chinese test records, their output becomes a typed configuration, and that configuration alone selects the runs, defines the input profile, checks the applicability of the trained bound model and sets the limits of the notching decision. This work makes four contributions. First, we define the task of driving a calibrated test analysis from text, with a schema of 76 fields, a gold configuration and outcome categories that separate identical results, rounding differences and silent errors, with escalations counted independently. Second, we place six deterministic validation gates between the LLM output and the numerical stage, covering units, physics, verbatim quotes, cross model agreement, the measurement exports and model applicability, and show that in this campaign all 20 LLM configurations then reproduce the gold notching decisions bit for bit while every real and injected error in a consumed field is stopped. Third, we evaluate the extraction itself with an error taxonomy, evidence grounding, cross model checking and repeat runs, which explains why each gate is needed and shows that hallucination detection requires more than quote checking alone. Fourth, we report a negative result of practical value. Direct LLM curve forecasting, in the style of LLMTime [11], does not exceed linear extrapolation, whereas a small physics-structured network with conformal calibration reduces the over-notched bandwidth by 86% to 89% at coverage $c = 0.90$, which delineates where direct LLM prediction should give way to a trained model.

## 2. Related Work

**Structured extraction with LLMs.** LLMs extract entities, relations and records from scientific text with a schema given in the prompt [1], generative information extraction has been surveyed in [2], and maintenance records of aircraft have been processed in this way [3]. Output formats can be enforced by grammar-constrained decoding [12], whereas API models usually offer only a JSON mode, and format restrictions can change model performance [13].

**LLMs that drive tools.** Interleaving reasoning with actions [4], learning to call tools [5] and delegating computation to an interpreter [6] let a language model control external programs. In chemistry, LLMs connected to expert tools planned syntheses [7] and ran laboratory experiments [8]. These systems give the model control of the workflow. We keep the workflow fixed and deterministic and let the model supply only its configuration, which is then checked.

**Hallucination and verification.** Sampling-based consistency checks detect hallucinated content without access to model internals [14], agreement across samples improves reasoning [15], and planned verification questions reduce hallucination [16]. LLMs are also used to produce annotations, with known risks for evaluation [17]. Our gates use verbatim quotes, a second model, measurement files and physical constraints.

**LLMs for numbers and calibrated surrogates.** LLMs can forecast numerical series written as text [11], but ablations found that removing the LLM often does not hurt [18]. Virtual shaker models predict sine test responses [19], and jointed structures such as hinged solar arrays show nonlinear, amplitude-dependent dynamics [20, 21]. Structural vibration responses have been predicted with neural networks and neural operators [22, 23], and conformal prediction gives coverage guarantees under exchangeability [24, 25] that weighted variants extend [26].

## 3. Task and Data

### 3.1 In-house dataset

We use an in-house laboratory test dataset from the acceptance-level sine vibration tests of the two wings, denoted +Y and −Y, of a stowed deployable solar array of a LEO satellite (Figure 1). The panels are clamped at hold-down points by hold-down and release mechanisms, whose release devices can be driven by shape memory alloy actuators [27]. Each wing was excited along three axes with a low-level sweep at 0.5 g before and after each acceptance-level sweep, and the response limit for notching was 65 g. The dataset contains the measured curves of the response locations, 22 on the +Y wing and 20 on the −Y wing, exported from the LMS test system (Siemens Simcenter Testlab, formerly LMS Test.Lab), and Chinese textual records of the test, hereafter called the test records. Transmissibility curves relative to the average of the four control channels are resampled onto 384 logarithmic points from 5 to 200 Hz for the low-level sweep and onto the first 312 of these points, spanning 5 to 100 Hz, for the acceptance level. This gives 498 pairs of low-level and acceptance-level curves, 198 from the +Y wing and 300 from the −Y wing, of which 378 come from complete runs. Appendix A gives details. The markers in Figure 1 follow the channel numbering of the LMS system, where number 15 is unused, and their placement is schematic rather than the exact sensor coordinates. The axis convention with Z normal to the panel plane is assumed for illustration.

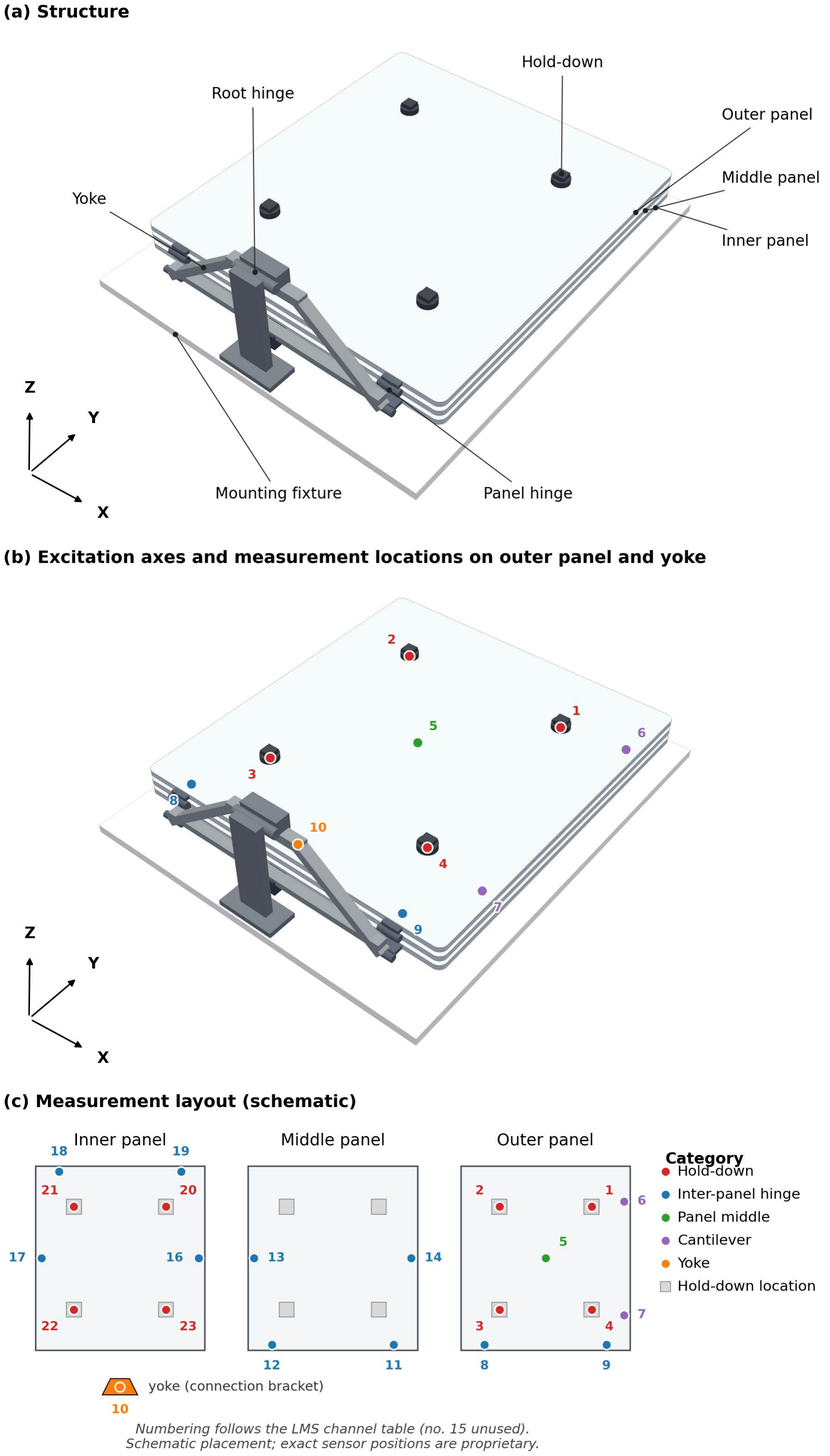

**Figure 1.** Simplified generic model of the stowed solar array wing. (a) Structure. (b) Excitation axes and the measurement locations on the outer panel and the yoke. (c) Schematic plan layout of the measurement points of one panel. The model is derived for illustration only. Dimensions, details and exact sensor positions differ from the tested hardware.

### 3.2 Task definition

Let D be the test records of a campaign, X the measurement exports and S a schema of 76 fields, of which 32 are numeric, 19 categorical and 25 lists. An extractor maps D to a configuration ĉ that assigns a value or null to every field. A subset R of the fields is consumed by the numerical stage, and the others, such as the axis order or the abort limit, are kept for audit. Gates assign every field f a decision $d_f(\hat{c})$, which is accepted, rejected or uncorroborated. The pipeline then returns

$$\hat{N} = \begin{cases} \Phi(\tau(\hat{c}), X) & \text{if } d_f(\hat{c}) = \text{accepted for all } f \in R \\ \text{STOP} & \text{otherwise} \end{cases} \quad (1)$$

where τ maps the configuration to typed values and Φ is the deterministic numerical stage that ends in notching decisions. A rejected audit field raises a flag but does not stop the run. With the gold configuration c* the reference decisions are N* = Φ(τ(c*), X). An LLM configuration is scored by its driving outcome, which is exact agreement if N̂ equals N* bit for bit, rounding agreement if all over-notch and under-notch totals differ by less than 0.05 octave and coverage by less than 0.01, and silent error if it runs with any larger difference. Independently of this score, a run that stops or flags a field for a human is an escalation, and a stop of a correct configuration is a false stop. The goal is no silent error at a low rate of false stops.

The extraction itself is scored against the gold. Before scoring, amplitudes are converted to g or to millimeter zero to peak, categorical values are mapped to canonical strings, and lists become sets. The atomic facts of a configuration are its non-null scalar fields and the elements of its non-null lists, 224 in the gold. A numeric field counts as correct within 1% of the gold value. A wrong scalar counts as one false positive and one false negative, list elements are matched individually, and a non-null value for a null gold field is a false positive. Precision, recall and F1 are computed from the pooled counts, and exact match is the fraction of fields whose normalized value equals the gold, null included.

# 4. LLM Extraction

### 4.1 Schema, gold and annotation rule

The schema has ten groups covering document metadata, the low-level sweep, the acceptance input segments, control, limits, tolerances, condition changes, runs per wing and axis, measurement points with their Chinese names and roles, and anomalies (Appendix B). Fields are filled only from explicit statements. In particular the abort limit is filled only if the records state a numeric abort threshold. The gold configuration was prepared with the help of a coding agent and passed 51 automated checks against the records and the measurement file metadata. The records state neither an abort threshold nor the limited points, so both are null in the gold, and the 36 limited channel entries of the campaign, one per limited point, wing and axis, exist only in the measurement files.

A second annotation was produced independently by a separate coding agent following a written guideline in which every non-null value needs a verbatim Chinese evidence span. The agent saw neither the gold nor model outputs before its annotation was saved and hashed, and all 64 evidence spans occur in the records. The two annotations agree on all 76 fields, with a Cohen's kappa of 1.0 on the 19 categorical fields. The agreement script counts 236 atomic facts because it counts each of the 12 null fields as one fact. Agreement between two agents does not replace human inter-annotator agreement; it shows that the gold is consistent with the written rules.

### 4.2 Models, inputs and baseline

All calls were made on 14 November 2026 through two endpoints that are aliases without a fixed snapshot. In a probe on that day the deepseek-chat endpoint of the DeepSeek API returned the server model field deepseek-flash, which the provider documentation maps to DeepSeek-V4.1-Flash-1017 in non-thinking mode [28, 29]. The 166 DeepSeek responses cached earlier on the same day did not record this field, so for them the model is inferred from the date and the documentation rather than observed. We refer to this model as DeepSeek V4.1 Flash. The qwen-plus endpoint of Alibaba Cloud Model Studio in the Beijing region is documented as equivalent to qwen-plus-2025-12-01 [30, 31]. It was used in non-thinking mode and is referred to as Qwen-plus. Asked for their own version, both models named versions that disagree with the documentation, and such answers were not used.

We used temperature 0 and JSON output mode, and the prompt contains the schema as a JSON template, the extraction rules and the text. Five inputs are compared. The curated excerpt contains the parts of the records that state test conditions, execution and sensor lists, and the chunked input splits it into four parts whose outputs are merged. The complete input contains all records. A layout perturbation of the excerpt changes table separators, date formats and headings, and a wording perturbation states the same facts in m/s², as peak-to-peak displacement and with paraphrased sentences. Each input is extracted in a plain mode and in an evidence mode, E1, in which every value carries a verbatim quote. A rule-based extractor with regular expressions, written for the curated excerpt, serves as the baseline. Prompts are given in Appendix B.

Figure 2 shows the pipeline. Of the 20 hard-coded conditions found in the earlier analysis code, 14 are now read from the LLM output, 2 of them for audit only, and 3 are derived from the measurement exports, such as the limited points and an anomalous channel. The remaining 2 are method parameters, such as the 20 g relevance threshold of the calibration, and 1 is an implementation detail. Appendix C lists them with the gates that check each one.

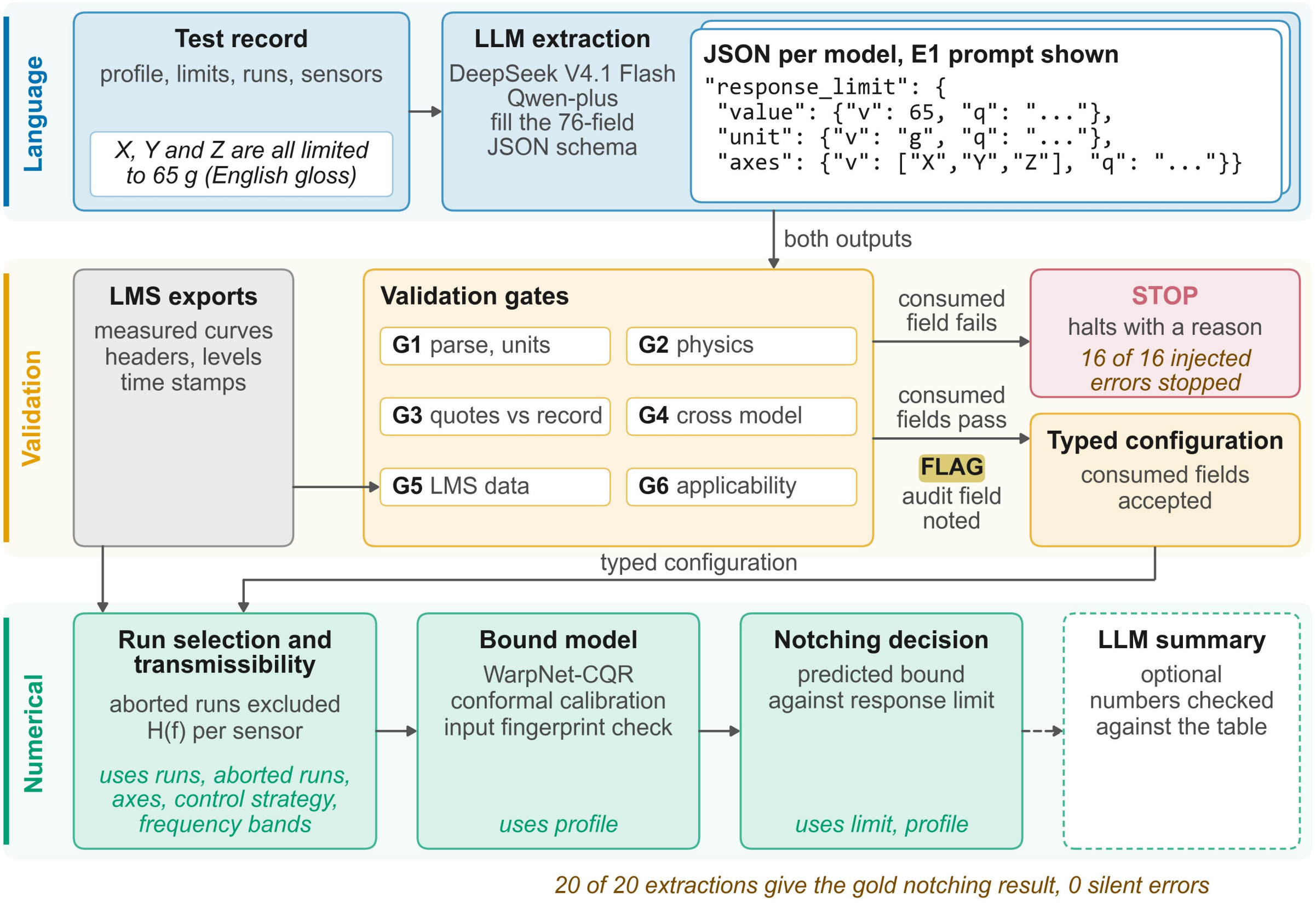

**Figure 2.** Overview of the pipeline in three lanes. In the language lane DeepSeek and Qwen each read the test record, shown here as an English gloss, and return a 76-field configuration. In the validation lane six gates check every field, a failed consumed field stops the run with a stated reason and a failed audit field is flagged. In the numerical lane the accepted configuration drives run selection, the calibrated bound model and the notching decision, and each box names the fields it consumes. The dashed LLM summary is optional and never enters the decision. Tags give the outcome of the final validation policy.

**Gates.** Gate G1 requires a parseable value with an explicit unit, converts m/s² to g and peak-to-peak to zero-to-peak displacement, and never substitutes a default. A converted value keeps only the precision of its source, so that 637.7 m/s² becomes 65 g rather than 65.005 g. G2 checks physical plausibility, such as a contiguous profile that is continuous in level, a response limit between 1 and 20 times the top input level and run lists that start and end with a low-level sweep. G3 applies the quote verifier of Section 6.2 to evidence-mode outputs. G4 compares the field with the extraction of the other model from the same input. G5 compares it with facts read from the measurement exports, such as the measured control spectrum against the profile, the limit marks, the number of acceptance files and the axis order from file timestamps. G5 thus checks consistency between the text and the exports that the numerical stage reads, not against independent ground truth. G6 checks that the bands and levels equal those for which the bound model was trained, within 2%. Appendix C lists all checks and tolerances.

# 5. Validation Gates and the LLM-Driven Pipeline

**Decision policies.** A failure of G1, G2, G5 or G6 rejects a field. Otherwise the field needs corroboration. The lenient policy accepts it if G3 or G4 passes. The consensus policy, our default, requires G4 to pass and uses G3 only when no second model is available, and a disagreement between the models is resolved in favor of a value that passes G5 and is not contradicted by its quote. The strict policy accepts a field only if neither G3 nor G4 fails.

**Repairs.** Two repairs act before the gates decide. Stage A extends the quote verifier to elided quotes. A quote written with an ellipsis passes only if every fragment occurs verbatim in the records and the fragments occur in order, and the value checks then run on the fragments alone. Stage B asks the same model once more for the consumed fields whose quotes still fail, with a template restricted to these fields and only the chunk of the records they came from. Both models are repaired before they are compared with each other. Gates and repairs never see the gold. Appendix C gives the re-ask prompt.

**Numerical stage.** The typed configuration selects the files of each wing and axis, orders acceptance runs by their timestamps, marks complete and aborted runs, computes transmissibility relative to the control channels, evaluates the input profile A(f) and builds the dataset. A fingerprint of everything the trained models saw is compared with the training artifact. The bound model of Section 7 is not retrained in routine runs. Its stored predictions are reused by channel, and a configuration whose fingerprint differs is marked as stale. Conformal calibration with virtual units and the notching decision are recomputed from the configuration. With the gold configuration the pipeline reproduced all 14 arrays of the hand-coded dataset and the coverage and over-notch of the 8 calibrated settings of Section 7 (2 directions, 2 levels and 2 calibration score sets) bit for bit, and all 20 LLM configurations rebuilt from cached responses equaled the saved outputs. To test the training stage as well, we retrained WarpNet-CQR, the conformalized quantile regression bound model of Section 7, with the same seeds on the dataset built from the gated configuration of DeepSeek V4.1 Flash on the curated excerpt. The dataset equaled the training artifact, and the predictions differed only by GPU nondeterminism, with a median of $5 \times 10^{-5}$ dex ($\log_{10}$ units) and a maximum of 0.013 dex. The over-notch at $c = 0.95$ summed over both directions changed from 88.93 to 89.04 octaves, with coverage 1.00 and no under-notch in both.

As an optional final step, an LLM writes a short notching summary from the per-axis table, with one clause per wing and axis. A checker assigns every number in the summary to its wing, axis and quantity and compares it with that cell of the table. All 36 numbers of each model were correctly attributed, and the checker detected all 30 misattributions injected into each summary by swapping numbers between cells, which a check of mere membership in the table missed. The summary is not used for any decision.

# 6. Experiments

### 6.1 Extraction accuracy and error taxonomy

**Table 1.** Extraction accuracy over 224 atomic facts in plain mode. Macro F1 averages the F1 of the ten field groups, with a group that has neither gold facts nor predictions counting as 1. File checks are nine consistency checks against the measurement files. The rule baseline was not run on the chunked input.

| Input | Extractor | Precision | Recall | F1 | Macro F1 | Exact match | File checks |
|---|---|---|---|---|---|---|---|
| Curated excerpt | DeepSeek V4.1 Flash | 1.000 | 1.000 | 1.000 | 1.000 | 1.000 | 9 of 9 |
| Curated excerpt | Qwen-plus | 0.991 | 0.996 | 0.993 | 0.979 | 0.974 | 8 of 9 |
| Curated excerpt | Rule baseline | 1.000 | 1.000 | 1.000 | 1.000 | 1.000 | 9 of 9 |
| Curated excerpt, four chunks | DeepSeek V4.1 Flash | 1.000 | 0.991 | 0.996 | 0.999 | 0.987 | 9 of 9 |
| Curated excerpt, four chunks | Qwen-plus | 0.991 | 0.969 | 0.980 | 0.977 | 0.947 | 8 of 9 |
| Complete records | DeepSeek V4.1 Flash | 1.000 | 1.000 | 1.000 | 1.000 | 1.000 | 9 of 9 |
| Complete records | Qwen-plus | 0.996 | 1.000 | 0.998 | 0.989 | 0.987 | 9 of 9 |
| Complete records | Rule baseline | 1.000 | 0.982 | 0.991 | 0.999 | 0.974 | 9 of 9 |
| Layout perturbation | DeepSeek V4.1 Flash | 1.000 | 1.000 | 1.000 | 1.000 | 1.000 | 9 of 9 |
| Layout perturbation | Qwen-plus | 0.978 | 1.000 | 0.989 | 0.962 | 0.974 | 9 of 9 |
| Layout perturbation | Rule baseline | 1.000 | 0.067 | 0.126 | 0.374 | 0.303 | 2 of 9 |

| Input | Extractor | Precision | Recall | F1 | Macro F1 | Exact match | File checks |
|---|---|---|---|---|---|---|---|
| Unit and wording perturbation | DeepSeek V4.1 Flash | 1.000 | 0.982 | 0.991 | 0.999 | 0.974 | 9 of 9 |
| Unit and wording perturbation | Qwen-plus | 0.991 | 0.996 | 0.993 | 0.979 | 0.974 | 8 of 9 |
| Unit and wording perturbation | Rule baseline | 0.985 | 0.906 | 0.944 | 0.615 | 0.776 | 6 of 9 |

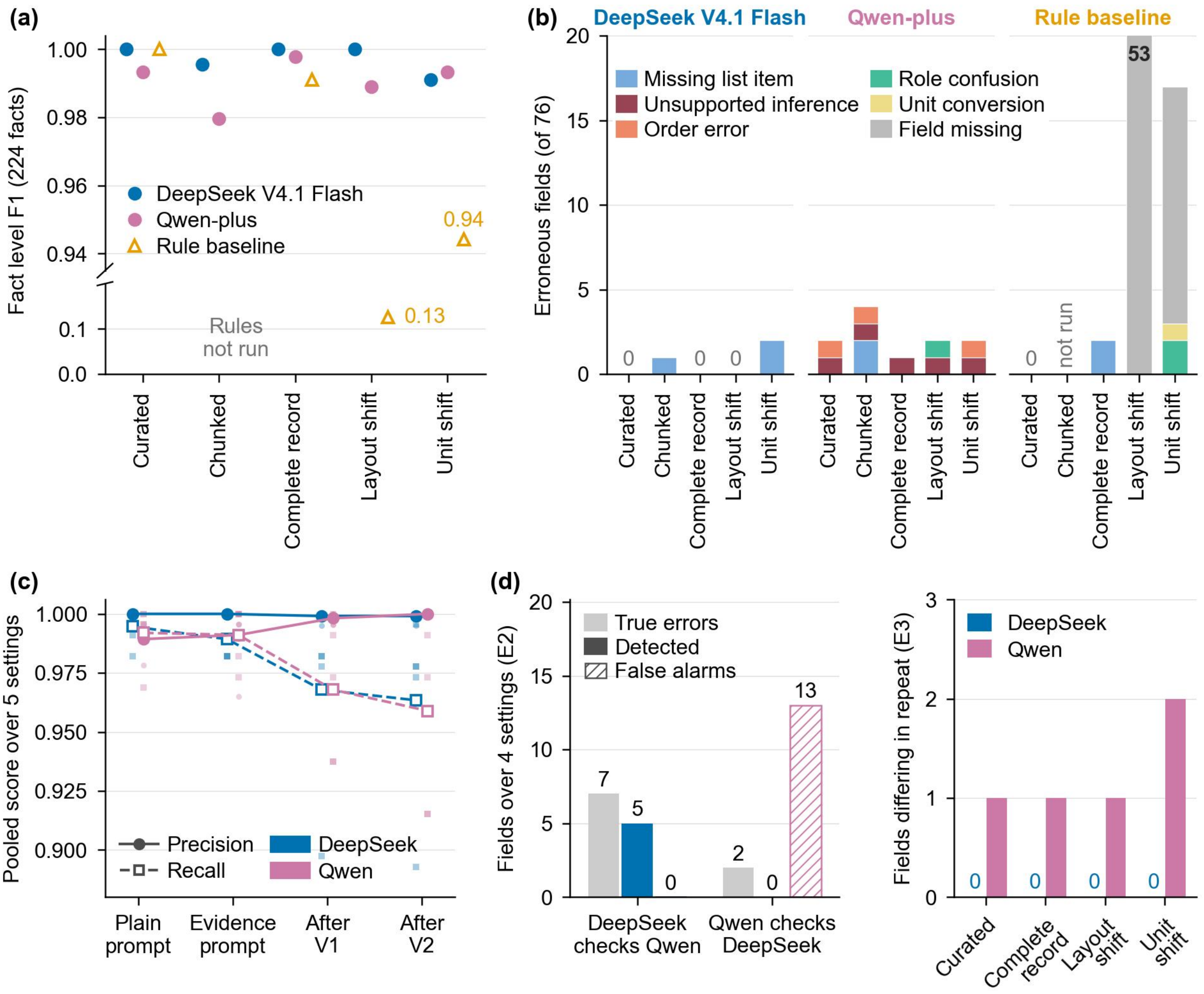

**Figure 3.** Extraction quality and verification. (a) Fact-level F1 over the 224 gold facts for each input setting. (b) Erroneous fields out of 76 by error type, the bar of the rule baseline under the layout shift is truncated. (c) Precision and recall of evidence grounding E1 pooled over the five settings, faint dots are single settings. (d) Left, cross model checking E2 of Section 6.3 over four settings. Right, repeat agreement E3 of Section 6.4 between two identical calls.

Table 1 and Figure 3 summarize the extraction. DeepSeek V4.1 Flash reached an F1 between 0.991 and 1.000 with a precision of 1.0 on all inputs, and Qwen-plus an F1 between 0.980 and 0.998 with a precision between 0.978 and 0.996. The rule baseline matched the LLMs on the text it was written for, but fell to 0.126 under the layout perturbation and 0.944 under the wording perturbation.

Table 2 classifies the errors. All errors of DeepSeek V4.1 Flash were missing elements of the sensor direction lists, an audit field. Qwen-plus made four kinds of errors. In all five inputs it treated the 65 g response limit as an abort threshold. The records state the 65 g limit and that limited points triggered emergency stops, so this is a plausible reading, but no numeric abort threshold is stated and the measurement files show abort lines at 81.83 g and 129.69 g. Under our annotation rule it is an unsupported inference, and a looser rule would change this count. In 3 inputs Qwen-plus gave the test order of the −Y wing axes as XYZ instead of XZY, and under the layout perturbation it listed the four control points as limited points, a role confusion. The rule baseline fails visibly by returning nothing, whereas the LLM errors are plausible values. The profile reconstructed from the extracted segments agreed with the hand-coded profile to machine precision, and to within 0.1% under the wording perturbation, where the conversion from m/s² rounds.

**Table 2.** Error taxonomy, given as the number of affected fields per model and input in plain mode. Unsupported inference denotes a value that is not explicitly stated, role confusion a point listed in a role it does not have.

| Model | Input | Missing list element | Unsupported inference | Order error | Role confusion | Unit conversion error | Other |
|---|---|---|---|---|---|---|---|
| DeepSeek V4.1 Flash | Curated excerpt | 0 | 0 | 0 | 0 | 0 | 0 |
| DeepSeek V4.1 Flash | Curated excerpt, four chunks | 1 | 0 | 0 | 0 | 0 | 0 |
| DeepSeek V4.1 Flash | Complete records | 0 | 0 | 0 | 0 | 0 | 0 |
| DeepSeek V4.1 Flash | Layout perturbation | 0 | 0 | 0 | 0 | 0 | 0 |
| DeepSeek V4.1 Flash | Unit and wording perturbation | 2 | 0 | 0 | 0 | 0 | 0 |
| Qwen-plus | Curated excerpt | 0 | 1 | 1 | 0 | 0 | 0 |
| Qwen-plus | Curated excerpt, four chunks | 2 | 1 | 1 | 0 | 0 | 0 |
| Qwen-plus | Complete records | 0 | 1 | 0 | 0 | 0 | 0 |
| Qwen-plus | Layout perturbation | 0 | 1 | 0 | 1 | 0 | 0 |
| Qwen-plus | Unit and wording perturbation | 0 | 1 | 1 | 0 | 0 | 0 |

### 6.2 Evidence grounding (E1)

To make every extracted value checkable, we repeated the extraction with an evidence prompt in which each non-null value must carry a verbatim quote from the test records. A deterministic verifier without any model then checks the quotes at two levels. Level V1 checks that every quote occurs in the input, ignoring whitespace, and that the value is consistent with its quote in its numbers, units and enumerated keywords and, for the axis order, in the positions of the quoted headings. Level V2 adds keyword rules for single fields, namely abort wording for the abort limit, limit wording for the response limit, the wing label for wing-scoped values and tolerance and spacing wording. A value that fails is set to null. The V2 rules were finalized after inspection of the errors of the evidence outputs, so their scores are optimistic for these records. Bugs in the verifier found during development were fixed without reference to the gold labels. Before verification, rule-based repairs removed 3 premature closing braces from the JSON of DeepSeek V4.1 Flash and moved 9 top-level groups that Qwen-plus had nested inside another group under the wording perturbation.

Table 3 gives the result. After V2 the precision of Qwen-plus was 1.0 on all five inputs. DeepSeek V4.1 Flash reached 1.0 on 4 inputs and 0.995 under the layout perturbation, where removing unsupported input segments left a wrong number of segments. The evidence prompt alone avoided the abort threshold of Qwen-plus in 3 inputs. In the curated excerpt and complete records, where the evidence output still contained it, V1 kept the value because the quoted sentence does state 65 g, and only the abort wording rule of V2 removed it. The evidence prompt also caused new errors. Under the wording perturbation Qwen-plus copied 4 accelerations in m/s² without converting them to g, which V1 removed. Recall dropped, to 0.915 for Qwen-plus and 0.893 for DeepSeek V4.1 Flash, because correct values carried abbreviated or incomplete quotes and were removed. Of the 61 fields that V2 set to null over all inputs, 53 had been correct and 8 wrong. The wrong axis order of the −Y wing was absent from all evidence outputs of Qwen-plus, but the repeat runs below show that this field changes between identical requests, so we do not credit the evidence prompt with the change. Quote verification thus turned the remaining plausible errors into missing values, which a downstream check can see, at a cost in recall.

**Table 3.** Evidence-grounded extraction over 224 atomic facts. Original is the extraction of Table 1, evidence the output of the evidence prompt before verification, and V1 and V2 the same output after the generic and the field-specific quote checks. The last column counts the fields that V2 set to null and how many of them had been correct.

| Model | Input | F1 original | F1 evidence | F1 V1 | F1 V2 | Precision V2 | Recall V2 | Correct among removed |
|---|---|---|---|---|---|---|---|---|
| DeepSeek V4.1 Flash | Curated excerpt | 1.000 | 0.991 | 0.991 | 0.989 | 1.000 | 0.978 | 1 of 1 |
| DeepSeek V4.1 Flash | Curated excerpt, four chunks | 0.996 | 0.991 | 0.989 | 0.986 | 1.000 | 0.973 | 1 of 2 |
| DeepSeek V4.1 Flash | Complete records | 1.000 | 1.000 | 1.000 | 0.998 | 1.000 | 0.996 | 1 of 1 |
| DeepSeek V4.1 Flash | Layout perturbation | 1.000 | 1.000 | 0.944 | 0.941 | 0.995 | 0.893 | 20 of 20 |
| DeepSeek V4.1 Flash | Unit and wording perturbation | 0.991 | 0.991 | 0.991 | 0.989 | 1.000 | 0.978 | 1 of 1 |
| Qwen-plus | Curated excerpt | 0.993 | 0.998 | 0.998 | 1.000 | 1.000 | 1.000 | 0 of 1 |
| Qwen-plus | Curated excerpt, four chunks | 0.980 | 0.986 | 0.968 | 0.956 | 1.000 | 0.915 | 12 of 12 |
| Qwen-plus | Complete records | 0.998 | 0.998 | 0.966 | 0.956 | 1.000 | 0.915 | 14 of 15 |
| Qwen-plus | Layout perturbation | 0.989 | 1.000 | 0.996 | 0.996 | 1.000 | 0.991 | 1 of 1 |
| Qwen-plus | Unit and wording perturbation | 0.993 | 0.974 | 0.986 | 0.986 | 1.000 | 0.973 | 2 of 7 |

### 6.3 Cross model checking (E2)

In a cross model check, each model received the records of the curated excerpt, complete records, layout perturbation and unit and wording perturbation together with the original output of the other model and was asked to flag unsupported or wrong fields. DeepSeek V4.1 Flash flagged 5 of the 7 known errors of Qwen-plus and raised no false alarm. Qwen-plus flagged 0 of the 2 known errors of DeepSeek V4.1 Flash and raised false alarms on 13 correct fields, in 3 inputs asking for the 65 g abort threshold that its own extraction had produced. These counts rest on nine known errors in total.

### 6.4 Repeat runs (E3)

To assess the stability of the outputs, the original extraction was repeated once with identical requests on the curated excerpt, complete records, layout perturbation and unit and wording perturbation. DeepSeek V4.1 Flash returned the same value in all 76 fields on every input, although the JSON text differed in 2 of the 4 inputs. Qwen-plus agreed on 74 or 75 of 76 fields, and the disagreements concerned the axis order of the −Y wing and the limited points of the −Y wing under Y-axis excitation. Temperature 0 therefore does not make the API outputs deterministic, and the error counts of Table 2 are samples from a single run rather than fixed properties of the models.

### 6.5 End-to-end driving outcomes and error propagation

**Table 4.** Driving outcomes of the 20 LLM configurations, two models, five inputs and two modes, for the successive pipeline versions. Identical means that every downstream metric equals the gold run bit for bit. Escalated counts runs that stopped or flagged at least one field for a human, and silent error counts runs with a wrong consumed value that did not stop. Versions v1 to v3 were developed in this order on these configurations.

| Pipeline version | Identical to gold | Escalated | Stopped | Silent error |
|---|---|---|---|---|
| v0, no gates | 16 | 0 | 0 | 1 |
| v1, gates, consensus | 16 | 8 | 1 | 0 |
| v1, gates, strict | 12 | 15 | 6 | 0 |
| v2, plus source precision | 19 | 8 | 1 | 0 |
| **v3, plus repairs A and B, consensus** | **20** | **8** | **0** | **0** |
| v3, plus repairs A and B, strict | 19 | 12 | 1 | 0 |

Table 4 and Figure 4 give the end-to-end result. With the final pipeline v3 and the consensus policy, all 20 LLM configurations produced notching decisions identical to the gold run bit for bit, and none produced a silent error. The 8 escalations were all flags on audit fields of Qwen-plus, which did not stop the run, namely its abort threshold in 7 configurations, its −Y axis order in 3 and a wrong limited-points list of the −Y wing in 2. Without gates, version v0 gave identical results in 16 configurations, 3 differed only by rounding and one ran with wrong consumed values.

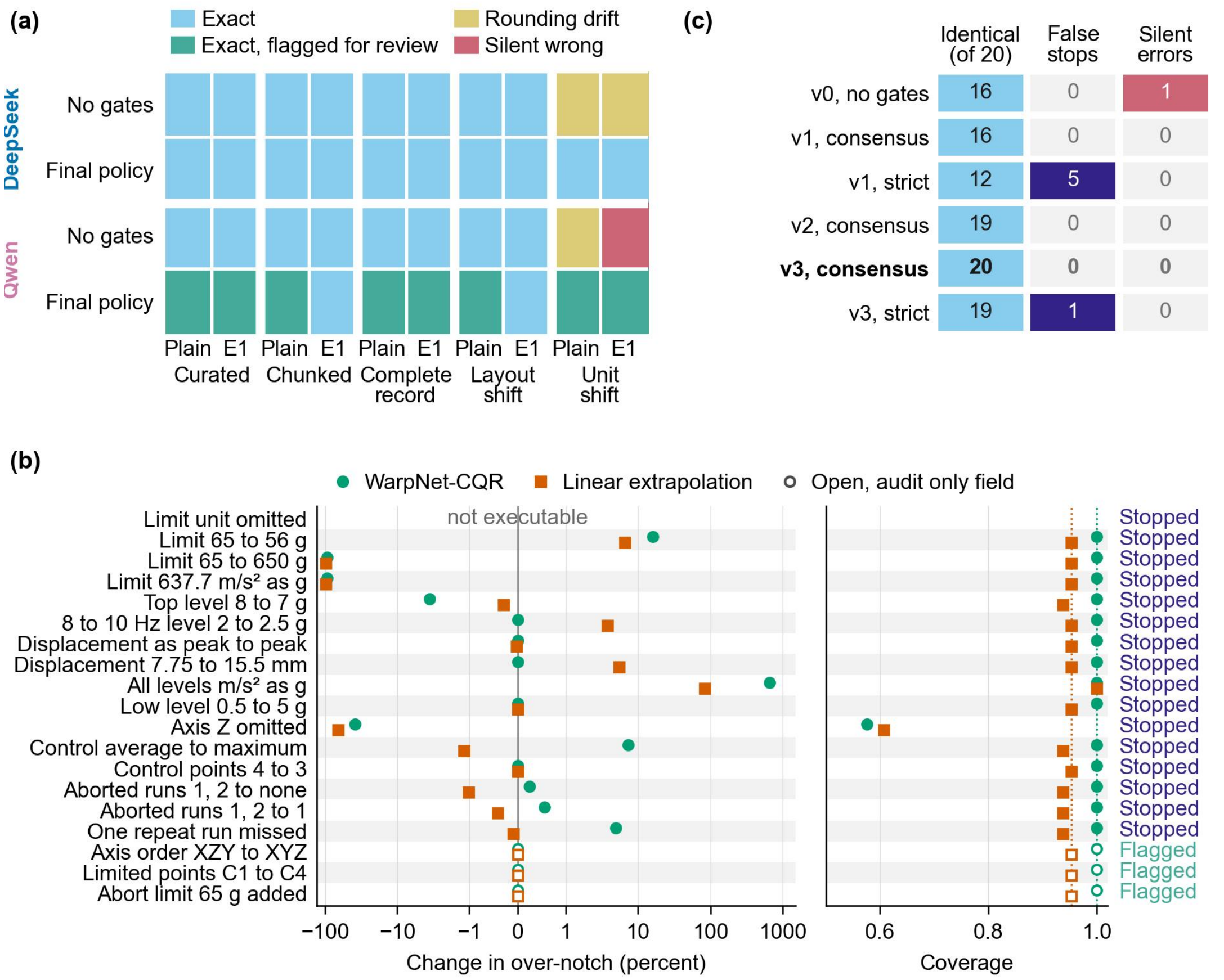

**Figure 4.** Downstream effect of extraction errors and pipeline versions. (a) Downstream outcome of the 20 extracted configurations without gates and under the final policy. (b) Change in over-notch and coverage at the 95 percent level for each of the 19 injected errors without gates. (c) Successive pipeline versions on the same 20 configurations, counting identical outcomes, false stops of correct configurations and silent errors.

That configuration was the evidence output of Qwen-plus on the wording perturbation before the re-ask. It copied accelerations given in m/s² and labeled them g, which inflates every level by the factor 9.81, so the profile levels became 19.6 g and 78.5 g instead of 2 g and 8 g, the low level 4.9 g and the limit 637.7 g. Gates G2, G3, G4, G5 and G6 all rejected the profile and the run stopped with these reasons. Without gates it would have completed with 21.3 more octaves of over-notch at $c = 0.95$, 24% above the gold run. This effect is computed with the stored predictions of WarpNet-CQR, because the fingerprint of the inflated dataset differs from the training artifact, and thresholds expressed in g, such as the 20 g relevance threshold of the calibration, then act on inflated values, so the figure is an estimate. The targeted re-ask of stage B then returned values in g that passed all gates, and the repaired configuration ran with identical results.

The versions show where each part of the result comes from. Gates alone, in v1, removed the silent error but left 3 configurations on the wording perturbation with small rounding differences, because values converted from m/s² carry rounding, such as 65.005 g for 65 g. We found this on a perturbed input that was meant to be held out, and the source precision rule of v2 was added subsequently, raising the number of bit identical configurations from 16 to 19. The repairs of v3 were designed on the curated and chunked inputs and frozen before they were run on the complete records and the two perturbations. On these held-out inputs they reduced the false stops of the strict policy among correct configurations from 4 to 1. Over all 20 configurations, development inputs included, the strict policy stopped 5 correct configurations in v1 and 1 in v3 (Figure 4c). Under the consensus policy the repairs changed no correct outcome but repaired the one wrong configuration.

### 6.6 Injected errors and gate policies

Real errors in consumed fields were rare, so we injected 19 errors into the evidence output of DeepSeek V4.1 Flash on the curated excerpt, keeping the original quotes, 16 in consumed fields and 3 in audit fields (Appendix C). They cover unit slips, decimal slips, wrong levels, the displacement convention, an omitted axis, the control scheme and the run list. Under the consensus policy all 16 consumed errors stopped the run and all 3 audit errors were flagged, with and without G5. Without gates, the pipeline could not run in 1 case, a limit without unit, and 2 errors left the results unchanged, but the remaining errors had large effects. A limit of 56 g instead of 65 g increased the over-notch of WarpNet-CQR at c = 0.95 by 16%. A decimal slip to 650 g left 1.3 octaves of true exceedance unprotected. Reading all levels in $m/s^2$ as g increased the over-notch by 663%, and omitting the Z axis lowered the coverage of relevant channels to 0.58. In total, 3 injected errors would have reduced protection.

**Table 5.** Corroboration policies. Lenient accepts a field if the quote check or the other model agrees, consensus requires agreement of the two models, resolved by the measurement exports when they disagree, and strict requires that neither check fails. The counts are over 23 correct configurations and 1 configuration with a wrong consumed value among the 20 final and 4 pre-re-ask outputs.

| Policy | Gate set | Correct configs stopped | Real consumed errors run | Injected consumed errors run | Injected audit errors flagged |
|---|---|---|---|---|---|
| Lenient | without G5 | 0 of 23 | 0 of 1 | 2 of 16 | 3 of 3 |
| Lenient | with G5 | 0 of 23 | 0 of 1 | 0 of 16 | 3 of 3 |
| Consensus | without G5 | 0 of 23 | 0 of 1 | 0 of 16 | 3 of 3 |
| Consensus | with G5 | 0 of 23 | 0 of 1 | 0 of 16 | 3 of 3 |
| Strict | without G5 | 4 of 23 | 0 of 1 | 0 of 16 | 3 of 3 |
| Strict | with G5 | 4 of 23 | 0 of 1 | 0 of 16 | 3 of 3 |

Table 5 compares the policies over the 20 final configurations and the 4 stored outputs before the re-ask. The lenient policy without G5 ran 2 injected errors, the omitted Z axis and a single aborted run instead of two. Both are subsets of what the records state, so their quotes are verbatim and pass G3, and only the second model disagreed. Quote checking alone therefore accepts errors of omission, and agreement between models is essential. The strict policy stopped every error but also 4 correct configurations, because the quote verifier rejected correct values whose quotes did not contain them. Of these, 3 were outputs before the re-ask, and one, the evidence output of DeepSeek V4.1 Flash on the layout perturbation, kept a failing quote after it. The consensus policy stopped every consumed error without a false stop, with and without G5.

## 7. Bound Model in Brief

The numerical stage needs an upper bound on the acceptance-level response of a new unit from its low-level sweep. Linear scaling is not enough because the structure softens. From 0.5 g to 8 g the matched resonance frequencies changed by a median of −1.5%, against −0.3% between the repeated low-level sweeps, and the peak transmissibility ratio had a median of 0.95 over 204 modes (Appendix A). Given the low-level curve $x(f) = \log_{10} T(f)$, a method returns bounds $u_q(f)$ on the acceptance-level curve $y(f)$, indexed by a scalar q, and a notch is placed wherever $10^{u(f)} A(f)$ exceeds the limit L. For channel n the relevant set is

$$\mathcal{F}_n = \{f \in \mathcal{M}_n | 10^{y_n(f)} A(f) > 20\ g\} \tag{2}$$

where $M_n$ is the set of unmasked bins. A relevant channel is covered if $u_q(f) \geq y_n(f)$ on its whole relevant set, 59 of 180 test channels for +Y→−Y and 32 of 198 for −Y→+Y. The cost is the over-notched bandwidth summed over all test channels, in octaves,

$$O = \sum_n \int_{\mathcal{M}_n} \mathbf{1}\left[10^{u_n(f)} A(f) > L \ \wedge \ 10^{y_n(f)} A(f) \leq L\right] d\log_2 f \tag{3}$$

We train on all pairs of one wing and test on the complete pairs of the other. WarpNet-CQR is a dilated convolutional network with FiLM conditioning [32] whose conformalized quantile regression (CQR) head may only warp the frequency axis of the measured curve, mostly toward lower frequencies, and mostly attenuate it,

$$\hat{y}(f) = x\left(f e^{d(f)}\right) + g(f) \tag{4}$$

with a bounded warp $\bar{d}$ and gain g, and a second output gives an upper quantile $\hat{y}^{+} \geq \hat{y}$. It is pretrained on simulated pairs, trained on real pairs and on virtual units that share one random nonlinearity per unit, and used as a five-member ensemble [33]. With ensemble mean $\bar{u}$ and spread σ of $\hat{y}^{+}$ and a floor $\sigma_0 = 0.02$ dex, the bound family and the conformal score of a relevant calibration channel [25] are

$$u_q(f) = \bar{u}(f) + q(\sigma(f) + \sigma_0),\ s_n = \max_{f \in \mathcal{F}_n} \frac{y_n(f) - \bar{u}_n(f)}{\sigma_n(f) + \sigma_0} \tag{5}$$

and $\hat{q}$ is a weighted quantile at level $c(1 + 1/N)$ of out-of-fold scores of the training wing and of scores of 30 virtual units. Mixing simulated and real scores departs from exchangeability [26]. Appendix D gives the simulator, head, loss and calibration.

**Table 6.** Total over-notched bandwidth in octaves of bounds calibrated on the training wing for nominal coverage c, with the achieved coverage of relevant test channels in parentheses. All variants are in Appendix E.

| Method | +Y→−Y c = 0.90 | +Y→−Y c = 0.95 | −Y→+Y c = 0.90 | −Y→+Y c = 0.95 |
|---|---|---|---|---|
| **WarpNet-CQR (proposed)** | **33.1 (1.00)** | **56.7 (1.00)** | **23.5 (1.00)** | **32.3 (1.00)** |
| WarpNet-CQR w/o VU aug. and smoothing | 47.4 (1.00) | 91.0 (1.00) | 24.3 (1.00) | 37.0 (1.00) |
| FiLM-CNN + CQR head | 52.0 (1.00) | 110.8 (1.00) | 26.8 (1.00) | 38.4 (1.00) |
| Modal-parametric (GBM) | 35.4 (1.00) | 65.5 (1.00) | 19.6 (1.00) | 28.6 (1.00) |
| FiLM-CNN (base) | 81.3 (0.98) | 159.8 (1.00) | 34.1 (1.00) | 61.1 (1.00) |
| Linear extrapolation | 235.8 (1.00) | 263.0 (1.00) | 214.7 (0.84) | 234.4 (0.91) |

Table 6 gives the deployable comparison. Calibrated on the training wing alone, WarpNet-CQR reached coverage not lower than nominal in both directions and over-notched 33.1 and 23.5 octaves at c = 0.90, against 235.8 and 214.7 for linear extrapolation with a margin calibrated the same way, which reached only 0.84 coverage in the −Y→+Y direction. This is a reduction of 86% to 89%. In the −Y→+Y direction the modal-parametric baseline was lower (19.6 octaves), and the calibration is conservative for all learned methods. Linear extrapolation needs a large margin because its error grows with the local slope of the low-level curve as resonances shift (Appendix E), which an amplitude margin cannot represent and a frequency warp can.

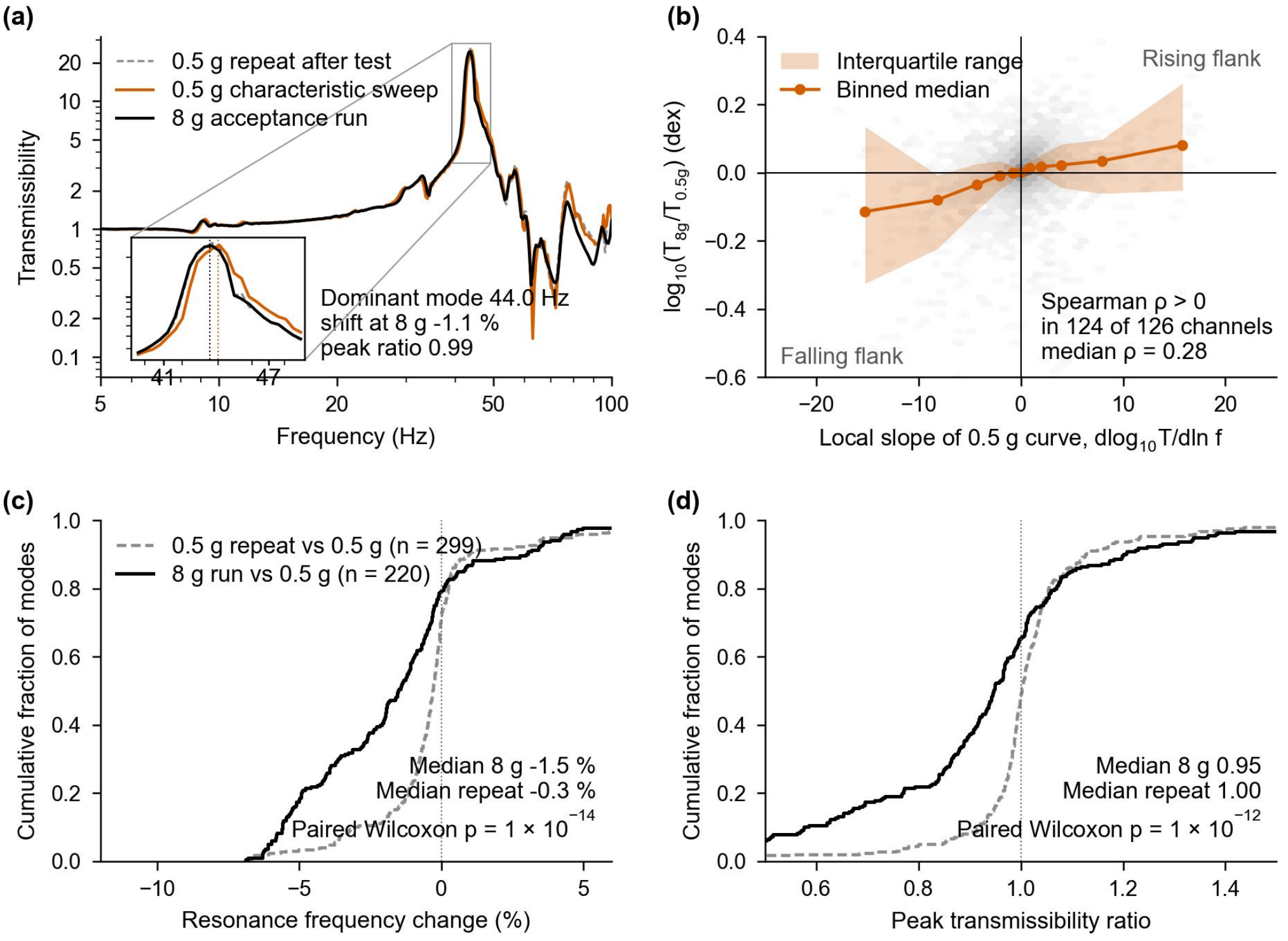

**Figure 5.** Level dependence of the measured transmissibility. (a) Example channel, chosen as the median dominant mode shift among the 20 in-axis channels whose dominant mode would exceed the 65 g limit at 8 g, the inset isolates the dominant mode. (b) Error of linear extrapolation against the local slope of the low-level curve. (c) Resonance frequency change of matched modes at 8 g against the post test repeat. (d) Peak transmissibility ratio at 8 g against the repeat.

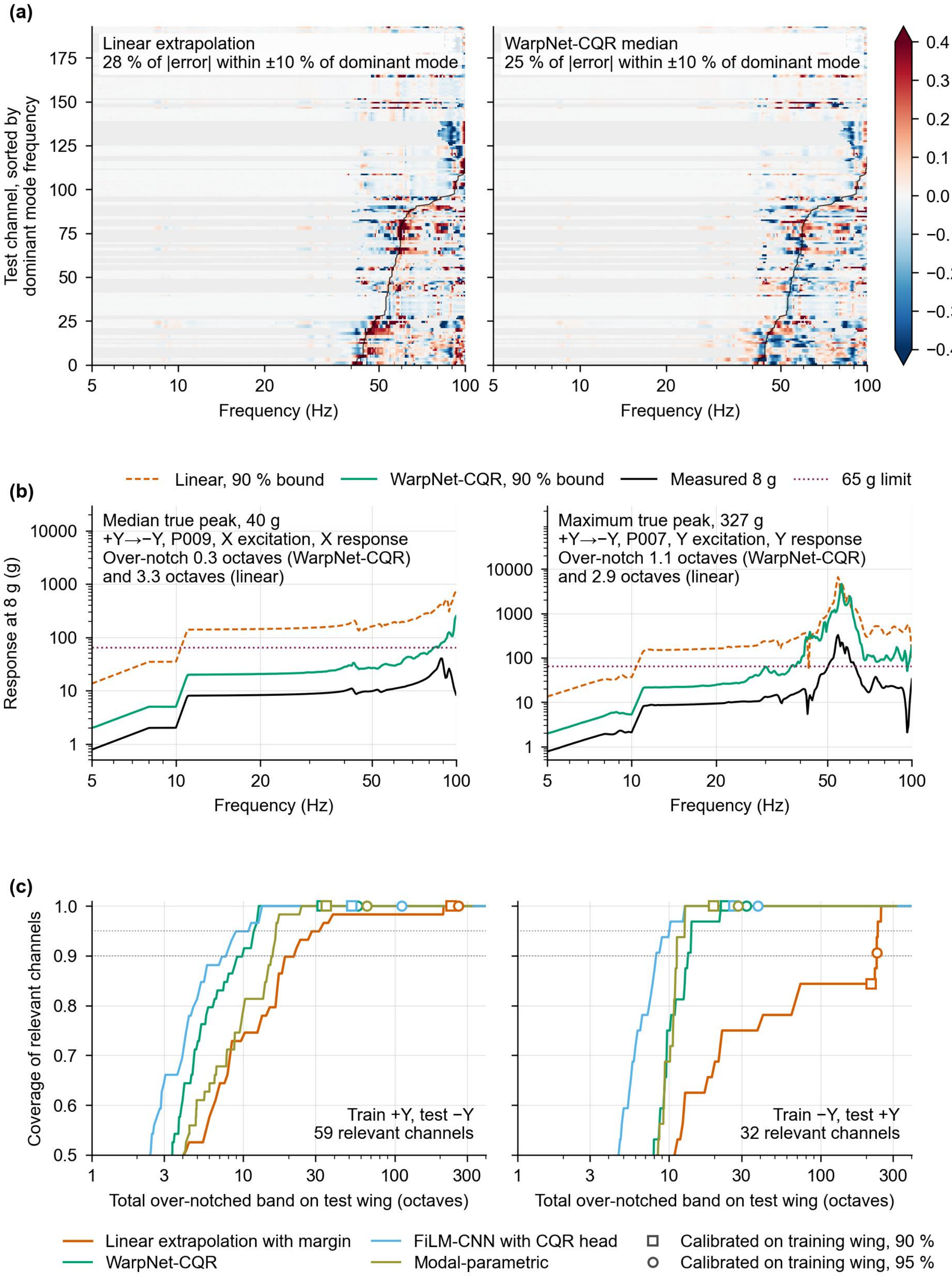

**Figure 6.** Prediction errors and calibrated bounds on the held out wing. (a) Point prediction minus truth in log10 transmissibility, channels sorted by the frequency of the dominant mode, grey marks masked bins and bins with a true transmissibility below 0.5. (b) Bounds calibrated on the training wing at 90 percent against the measured response of the relevant test channels with the median (40 g) and the maximum (327 g) true peak. (c) Coverage of relevant test channels against total over-notched bandwidth when the bound parameter is swept on the test wing, markers are the calibrated operating points.

Figure 5 shows the measured level dependence behind this design, the softening of the resonances and the growth of the linear extrapolation error with the local slope of the low-level curve. The example channel is P013 of the −Y wing under Y excitation, whose dominant mode at 44.0 Hz moves down by 1.1 percent with a peak ratio of 0.99 while the post test repeat stays on the original curve.

Bins within ±10 percent of the dominant mode are 7 percent of the evaluated bins but carry 28 percent of the absolute error of linear extrapolation and 25 percent of that of the WarpNet-CQR median. The total absolute error of the WarpNet-CQR median is 86 percent of that of linear extrapolation. Near the dominant mode the per channel mean absolute error is lower for WarpNet-CQR in 112 of the 165 test channels with at least 3 evaluated bins in that window (Wilcoxon $p = 2 \times 10^{-8}$). Figure 6 also shows the trade-off when the bound parameter is chosen on the test wing itself. Along this oracle frontier FiLM-CNN with a CQR head is slightly better than WarpNet-CQR (Appendix E), so the advantage of WarpNet-CQR lies at the deployable calibrated point.

## 8. LLM Numeric Forecasting Probe

A further question is whether the LLMs could also replace the bound model. We prompted both in the style of LLMTime [11]. Each curve was resampled to 48 log-spaced frequencies from 5 to 100 Hz and rendered with two decimal places. With $k = 0$ the prompt contains only the query low-level curve, and with $k = 4$ it adds the four training-wing pairs whose low-level curves are closest. The answer is a list of 48 values, interpolated back to the evaluation grid. The test channels are the first 30 relevant channels of each test wing in index order, which in the +Y→−Y direction all come from X-axis excitation. DeepSeek V4.1 Flash answered all 120 requests and Qwen-plus failed in 5 of 120, all at $k = 4$ with 47 or 49 values, so methods are compared on the 26 and 29 channels answered by both.

**Table 7.** Point prediction error on the 26 and 29 test channels answered by both LLMs, over unmasked bins with a true transmissibility of at least 0.5 and, in the relevant band columns, over bins with a true response above 20 g.

| Method | logRMSE +Y→−Y | logRMSE −Y→+Y | Relevant band +Y→−Y | Relevant band −Y→+Y |
|---|---|---|---|---|
| DeepSeek V4.1 Flash in context, k = 0 | 0.104 | 0.187 | 0.204 | 0.230 |
| DeepSeek V4.1 Flash in context, k = 4 | 0.109 | 0.216 | 0.256 | 0.312 |
| Qwen-plus in context, k = 0 | 0.106 | 0.193 | 0.198 | 0.225 |
| Qwen-plus in context, k = 4 | 0.270 | 0.314 | 0.623 | 0.431 |
| Neighbour gain transfer, k = 4, no LLM | 0.116 | 0.205 | 0.228 | 0.271 |
| Linear extrapolation, 48 points | 0.102 | 0.186 | 0.209 | 0.233 |
| Linear extrapolation, 312 points | 0.120 | 0.212 | 0.263 | 0.269 |
| WarpNet-CQR w/o VU aug. and smoothing | 0.083 | 0.159 | 0.181 | 0.202 |
| **WarpNet-CQR (proposed)** | **0.085** | **0.155** | **0.201** | **0.197** |
| 48-point resampling of the truth | 0.041 | 0.069 | 0.106 | 0.107 |

Table 7 and Figure 7 show the result. Without examples both LLMs essentially copied the low-level curve, with a pooled logRMSE of 0.104 and 0.187 for DeepSeek V4.1 Flash and 0.106 and 0.193 for Qwen-plus in the two directions, close to linear extrapolation on the same 48 points (0.102 and 0.186). With four examples DeepSeek V4.1 Flash became slightly worse (0.109 and 0.216), close to a gain transfer from the same neighbors without any LLM, and Qwen-plus much worse (0.270 and 0.314), driven by two channels in each direction with an error above 1 dex. WarpNet-CQR reached 0.085 and 0.155. Part of the advantage of the 48-point methods over linear extrapolation on the full grid is smoothing by the coarse resampling.

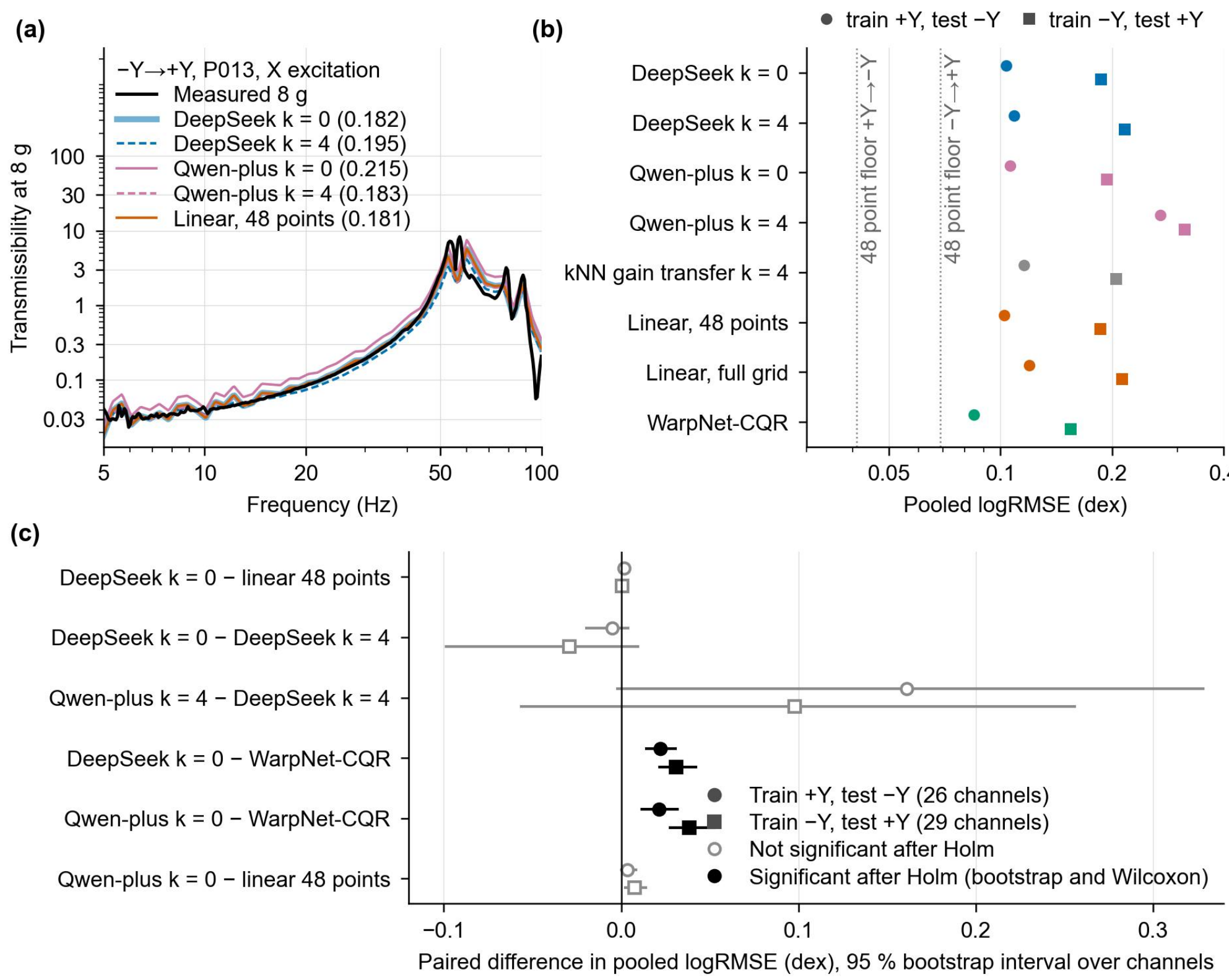

**Figure 7.** In context numeric forecasting by two LLMs on the 26 and 29 test channels answered by all runs of both models. (a) Example channel, the one with the median per channel logRMSE of DeepSeek V4.1 Flash at k = 0, LLM answers on 48 log spaced points are interpolated to the evaluation grid and per channel logRMSE values are in the legend. (b) Pooled logRMSE by condition, circles train +Y and test −Y, squares the reverse, vertical lines the error of the 48-point resampling alone. (c) Paired differences of pooled logRMSE with 95 percent bootstrap intervals over channels, filled markers survive the Holm correction for both the bootstrap and the Wilcoxon signed rank test.

In paired tests over channels with Holm correction [34] (Appendix F), only the gaps between the LLMs without examples and WarpNet-CQR were significant at the 5% level, in both directions, with differences of 0.021 to 0.039 dex. The LLMs did not differ significantly from linear extrapolation on the same 48 points.

# 9. Discussion

The LLMs were reliable drivers here because their role was narrow and every value they supplied was checked. They read the conditions, a fixed program used them, and gates stood between the two. The gates are cheap and deterministic, and each one catches a different class of error. Units and physics catch slips of scale, quotes catch values that are not in the text, a second model catches omissions whose quotes are still verbatim, the measurement exports catch values that contradict what was done, and applicability catches conditions the trained model has never seen. No single gate stopped all injected errors. Where a quote failed, asking the same model again with only the relevant chunk was cheaper than stopping, and it repaired the one wrong configuration. The extraction errors that remained were few and mostly plausible, which is the kind of error that a downstream program cannot detect by itself.

DeepSeek V4.1 Flash as checker found most errors of Qwen-plus, whereas the reverse check found none and raised many false alarms, so a second model is useful once its own reliability is known. In the gates we therefore compare the two extractions deterministically instead of asking one model to judge the other. For numbers the picture is reversed. The LLMs added nothing to linear extrapolation, whereas a small network whose output space encodes a shift and attenuation of measured resonances reduced over-notching at calibrated coverage.

The pipeline relies on the accelerometers of the test setup. Triboelectric sensors have been built into the joints of a snake-like spine mechanism [35] and into a sensor array for astronaut monitoring [36], and AI-enhanced BioMEMS sensors with sustainable energy harvesting have been reviewed as a route toward continuous monitoring and predictive analytics [41]. Thermoelectric generators that recycle waste heat from flexible displays offer a complementary power source for such sensor nodes [42]. Extending such self-powered sensors to hinges and hold-down points during batch testing is a natural next step.

For reference, Appendix G collects the counts used throughout this paper and explains how they relate, and Appendix H defines the terminology in one table.

## 10. Limitations

The evaluation rests on one complete acceptance test campaign of a single product with two wings, together with its test records, a gold configuration prepared with the help of a coding agent, and synthetic perturbations. The second annotation was also made by a coding agent. The annotation rule for inferred values decides the abort limit errors, the field-specific quote rules were written after inspecting the errors, the cross model check rests on nine known errors, and repeat runs show that single-run error counts vary. Both API endpoints are aliases whose models may change, so results are reproducible only through the cached responses.

The gate study has specific limits. The gates, the rounding rule and the repairs were developed and evaluated on the same campaign, and the repairs had a held-out split. Routine runs reuse stored predictions of the bound model, and retraining was tested for one configuration, so the downstream effects of errors that change the dataset are computed with a frozen model. Gate G5 reads the same measurement exports that the numerical stage uses, so it is a consistency check between text and data rather than independent ground truth. The injected errors were designed by us and cover the consumed fields, without claiming to enumerate every possible error. The trained model fixes the level ratio from 0.5 g to 8 g through its pretraining simulation, so G6 stops any campaign with other levels instead of adapting to it.

The numerical study has 59 and 32 relevant test channels in two cross-wing directions. The proposed configuration was selected after the benchmark, hyperparameters were set without a separate validation split, the virtual-unit prior is deliberately conservative, the mixed calibration is not covered by the exchangeability guarantee, and within notched bands the measured response is itself affected by notching. The warp head allows downward shifts of up to 0.15 in natural log frequency but upward shifts of only about 3%, while 12 of the 66 resonant channels shift upward by more, by up to 11%. The in context comparison uses one prompt format and channels from one excitation axis in one direction.

## 11. Conclusion

We let two LLMs drive a calibrated vibration test analysis from Chinese test records. They extracted a 76-field configuration with an F1 of 0.980 or higher, and six deterministic gates decided which values the run selection, transmissibility, bound model, calibration and notching decision could use. With all gates and the consensus policy, all 20 LLM configurations reproduced the gold notching decisions bit for bit and none produced a silent error, while the one real and all 16 injected errors in consumed fields were stopped. Retraining the bound model from an LLM configuration reproduced its results within GPU nondeterminism. Quote checks alone were not enough, and agreement between two models was needed. Asked to forecast the response directly, the LLMs reproduced linear extrapolation, while a calibrated physics-structured network reduced the over-notched bandwidth by 86% to 89% at $c = 0.90$. In this campaign the LLMs reliably drove the analysis because they supplied the configuration and deterministic code checked it and made the decision. Records from other organizations and further units are needed to confirm this.

## Reproducibility

LLM responses are cached and the pipeline refuses uncached prompts unless explicitly allowed. WarpNet-CQR has 178,308 parameters per ensemble member. Models were trained on one NVIDIA RTX 5090 laptop GPU with 24 GB memory, and training and calibration of one method in one direction took about 3 to 17 minutes. The pipeline with all 20 configurations and 19 injections runs in minutes on a CPU. The simulator uses seed 0, the ensemble members seeds 100 to 104, the virtual units seed 1 and the bootstrap seed 0. Code, prompts, schema, gold configuration and cached responses can be shared on request, subject to the confidentiality of the dataset.

## Ethics Statement

The dataset use for publication requires the consent of the data owner. The satellite designation and organizations are not named. The test records were sent to two public LLM APIs operated by companies located in China. The curated inputs contained no personal data. An open-weight model on local hardware avoids transferring records outside the organization. The gates reduce but do not remove the risk of a wrong test decision, and a test engineer remains responsible for the notch profile.

## AI Use Disclosure

Coding agents based on LLMs were used to write the analysis and pipeline code, to prepare the gold configuration under the direction of the authors, to produce the second annotation of Section 4.1, to assist in the analysis of the results. All results were produced by code in the repository of the authors, and the authors checked the text against the result files and take full responsibility for the content. The extraction, verification and forecasting experiments used 341 requests and 812,923 tokens, 415,298 for DeepSeek V4.1 Flash and 397,625 for Qwen-plus, and the pipeline stage added 10 requests, for a total of 843,374 tokens. The prices are those documented on 14 November 2026. For DeepSeek the off-peak price of that Saturday applies and for Qwen-plus the price of the Beijing region. Prices are converted at 7.1 CNY per USD and cached input is billed at 20% of the input price. The full set of calls cost about 0.09 USD for DeepSeek V4.1 Flash and 0.06 USD for Qwen-plus. A few Qwen-plus requests of a first attempt timed out and were not logged, at most about 8 requests and less than 0.03 USD.

## Data Availability

The in-house laboratory test dataset can be shared by the corresponding author on reasonable request with the consent of the data owner.

## Acknowledgements

To be added.

# Appendix A. Dataset Details

Each wing was excited along three orthogonal axes under four-point average control. The acceptance input was 7.75 mm displacement zero to peak from 5 to 8 Hz, 2 g from 8 to 10 Hz, a transition from 10 to 11 Hz and 8 g from 11 to 100 Hz. For every response channel the transmissibility is

$$T(f) = \frac{4|a_r(f)|}{\sum_{i=1}^{4} |a_{c,i}(f)|} \tag{6}$$

where $a_r$ is the response amplitude and $a_{c,i}$ are the amplitudes of the four control channels. Bins where the acceptance-level curve is undefined are masked. Raw acceptance-level peak accelerations are not used for evaluation, because notching during the runs depresses them, and all comparisons use transmissibility. For the mode statistics of Section 7, peaks of the low-level curve with a predicted acceptance-level response of at least 10 g between 8 and 120 Hz were matched within a symmetric window of 8% in frequency, the peak was refined by a parabola in log frequency, and matches at the window edge were rejected.

The matched resonance frequencies changed by a median of −1.5% (interquartile range −4.1% to −0.2%) from 0.5 g to 8 g, against −0.3% (−0.9% to +0.1%) between the two low-level sweeps, and the peak transmissibility ratio had a median of 0.95 (0.84 to 1.04) against 1.00. Paired Wilcoxon tests over 204 modes give $p = 1 \times 10^{-14}$ for the shift and $p = 1 \times 10^{-12}$ for the ratio. The resonances thus soften and lose amplitude at the acceptance level, but the size of the shift shows no trend with the predicted peak response (Spearman $\rho = -0.08$, $p = 0.22$), so the softening is not shown to be amplitude dependent.

# Appendix B. Prompts and Extraction Details

**Extraction system prompt.** The system prompt was used verbatim for both models.

```
You extract a machine-readable test configuration from a Chinese sine vibration test report of a satellite
solar array with two wings (+Y 翼 and -Y 翼). Return one JSON object that follows the template exactly (same keys,
same nesting). Rules: use only facts stated in the text; use null for any value or list that the text does not
state, never guess; numbers as plain numbers in the units named by the key; frequencies in Hz; for
displacement amplitudes keep the value as written and state the amplitude convention in unit (0-P means
0-peak); ignore tolerances signs (store magnitudes); runs lists the runs of one wing and axis in execution
order, one entry per run, using characteristic_pre, acceptance_sine or characteristic_post; aborted_runs and
anomalies.run use the 1-based index among the acceptance_sine runs of that axis; axis_order is the order in
which the excitation axes were tested for that wing; points lists every row of the sensor channel table of
that wing with the measurement point id (C1..C4 or the number) and the Chinese description copied verbatim;
limited_points are measurement point ids explicitly said to be response limited for that axis.
```

**JSON template.** The user message contains this template followed by the text in delimiters.

```
{
 "document": {
  "test_standard": "str",
  "test_dates": [
   "YYYY-MM-DD"
  ],
  "environment": [
   {
    "date": "YYYY-MM-DD",
    "temperature_C": 0.0,
    "humidity_RH": 0.0
   }
  ]
 },
 "characteristic_sweep": {
  "f_min_Hz": 0,
  "f_max_Hz": 0,
  "amplitude": 0,
  "amplitude_unit": "g",
  "sweep_rate_oct_per_min": 0,
  "performed_before_and_after_each_sine": true
 },
 "acceptance_sine": {
  "segments": [
   {
    "f_min_Hz": 0,
    "f_max_Hz": 0,
    "kind": "displacement|acceleration|transition",
    "value": 0,
    "unit": "e.g. mm 0-peak, mm peak-peak, g"
   }
  ],
  "axes": [
   "X"
  ],
  "sweep_rate_oct_per_min": 0,
```

```
  "n_sweeps": 0
 },
 "control": {
  "n_control_points": 0,
  "strategy": "average|maximum|minimum|single",
  "spacing_deg": 0
 },
 "response_limit": {
  "value": 0,
  "unit": "g",
  "axes": [
   "X"
  ]
 },
 "abort_limit": {
  "value": 0,
  "unit": "g"
 },
 "tolerances": {
  "duration_pct_min": 0,
  "duration_pct_max": 0,
  "frequency_pct": 0,
  "sine_amplitude_pct": 0
 },
 "condition_change": {
  "noted": true,
  "axes": [
   "X"
  ]
 },
 "wings": {
  "+Y": {
   "axis_order": [
    "X"
   ],
   "axes": {
    "X": {
     "runs": [
      "characteristic_pre|acceptance_sine|characteristic_post"
     ],
     "aborted_runs": [
      1
     ],
     "limited_points": [
      "point id"
     ]
    }
   },
   "sensor_directions": [
    "X"
   ],
   "points": [
    {
     "id": "C1 or 1",
     "name_cn": "verbatim Chinese name",
     "role": "control|response"
    }
   ]
  },
  "-Y": "same structure as +Y"
 },
 "anomalies": [
  {
   "wing": "+Y|-Y",
   "axis": "X|Y|Z",
   "run": 1,
   "type": "abort_limit_triggered|other"
  }
 ]
}
```

**Chunk merge rule.** In the chunked input each part is extracted separately with an added instruction to fill only the fields supported by that part. The outputs are merged recursively, dictionaries key by key, and for scalars and lists the first non-empty value in the order of the parts is kept.

**Curated excerpt.** The excerpt was cut from the records with regular expressions written for their section headings, and the two sensor tables were compacted to one row per sensor. This step is specific to these records and would need to be redone for records with another structure. The perturbed inputs were derived from the excerpt by fixed rewriting rules.

**In context prompt.** The system prompt and the user message of one k = 4 query are reproduced below, with the numbers abbreviated after the first ten values of each curve.

```
You are a numerical forecaster. Each item is one accelerometer channel of a satellite solar array wing in a
sine vibration test. 'low' is the log10 transmissibility (response divided by input) measured in a 0.5 g sweep
and 'high' is the log10 transmissibility of the same channel at the 8 g acceptance level. Both are sampled at
the same 48 log-spaced frequencies from 5 to 100 Hz. Predict 'high' for the query channel. Reply with a JSON
object {"high": [48 numbers with 2 decimals]} and nothing else.

Frequencies (Hz): 5.00,5.33,5.68,6.05,6.45,6.88,7.33,7.81,8.33,8.87, ... (48 values)

Example 1
low: 0.00,0.00,0.00,0.01,0.00,-0.01,0.01,0.00,0.00,-0.00, ... (48 values)
high: 0.00,0.01,0.00,-0.00,0.00,0.00,0.00,0.00,0.00,0.00, ... (48 values)

Example 2
low: -0.00,-0.00,-0.00,-0.00,-0.00,-0.00,-0.00,-0.00,-0.00,-0.00, ... (48 values)
high: -0.00,-0.00,-0.00,-0.00,-0.00,-0.00,-0.00,-0.00,-0.00,-0.00, ... (48 values)

Example 3
low: -0.00,0.00,0.00,-0.00,0.00,0.00,0.00,0.00,0.00,0.00, ... (48 values)
high: -0.00,-0.00,0.00,0.00,0.00,0.00,0.00,0.00,0.00,0.00, ... (48 values)

Example 4
low: -0.00,-0.00,0.00,0.00,0.00,0.00,0.00,0.00,0.00,0.00, ... (48 values)
high: 0.00,-0.00,0.00,0.00,0.00,0.00,0.00,0.00,0.00,0.00, ... (48 values)

Query
low: 0.01,-0.00,-0.01,0.00,0.01,0.01,-0.00,0.01,-0.01,-0.00, ... (48 values)
high:
```

# Appendix C. Pipeline Details

**Configurations.** Table 8 lists the outcome of every LLM configuration in the final pipeline under the consensus policy with all gates, including the stored outputs before the re-ask. Flagged fields are audit fields that were rejected and shown to the engineer without stopping the run.

**Table 8.** Outcomes of the 20 final configurations and the 4 outputs before the re-ask, consensus policy with G1 to G6.

| Configuration | Outcome | Identical to gold | Flagged audit fields |
|---|---|---|---|
| DeepSeek Curated plain | RUN | yes | none |
| DeepSeek Curated E1 | RUN | yes | none |
| DeepSeek Complete record plain | RUN | yes | none |
| DeepSeek Complete record E1 | RUN | yes | none |
| DeepSeek Chunked plain | RUN | yes | none |
| DeepSeek Chunked E1 | RUN | yes | none |
| DeepSeek Layout shift plain | RUN | yes | none |
| DeepSeek Layout shift E1 | RUN | yes | none |
| DeepSeek Layout shift E1 before re-ask | RUN | yes | none |
| DeepSeek Unit shift plain | RUN | yes | none |
| DeepSeek Unit shift E1 | RUN | yes | none |
| Qwen Curated plain | RUN | yes | abort limit, −Y axis order |
| Qwen Curated E1 | RUN | yes | abort limit |
| Qwen Complete record plain | RUN | yes | abort limit |
| Qwen Complete record E1 | RUN | yes | abort limit |
| Qwen Complete record E1 before re-ask | RUN | yes | abort limit |
| Qwen Chunked plain | RUN | yes | abort limit, −Y axis order |
| Qwen Chunked E1 | RUN | yes | none |
| Qwen Layout shift plain | RUN | yes | abort limit, −Y limited points |
| Qwen Layout shift E1 | RUN | yes | none |
| Qwen Layout shift E1 before re-ask | RUN | yes | none |

| Configuration | Outcome | Identical to gold | Flagged audit fields |
|---|---|---|---|
| Qwen Unit shift plain | RUN | yes | abort limit, −Y axis order |
| Qwen Unit shift E1 | RUN | yes | −Y limited points |
| Qwen Unit shift E1 before re-ask | STOP | no | −Y limited points |

**Hard-coded conditions.** Table 9 lists the conditions that the earlier analysis code fixed by hand, where each comes from in the LLM-driven pipeline and which gates check it.

**Table 9.** Audit of the hard-coded conditions of the earlier analysis code.

| Id | Condition | Source in the pipeline | Gates |
|---|---|---|---|
| H1 | acceptance input profile segments (5-8 Hz displacement / 8-10 Hz 2 g / 10-11 Hz log ramp / 11-100 Hz 8 g) | LLM acceptance_sine.segments | G1 units, G2 continuity and contiguity, G3 V2 quotes, G4 cross model, G5 control spectrum of complete acceptance runs |
| H2 | displacement unit and convention (7.75 mm 0-peak, g = 9.81 m/s²) | LLM segment unit (mm 0-peak or peak-peak, m/s² to g) normalized by schema.amp_canonical | G1, G2 continuity at 8 Hz |
| H3 | profile stored on the 256-point output grid then re-interpolated | PipelineConfig.profile() evaluated on the same two grids | regression test |
| H4 | response limit 65 g | LLM response_limit.value (+ unit) | G1, G2 limit above max input, G3, G4, G5 xlsx Spectrum Reference level |
| H5 | abort limit | LLM abort_limit (optional) | G2 abort above response limit, G4, G5 xlsx second Mark level |
| H6 | low-level (characteristic) band 5-200 Hz and input grid | LLM characteristic_sweep.f_min/f_max | G5 xlsx X axis range of characteristic runs, model applicability (must equal training band) |
| H7 | acceptance band 5-100 Hz and output grid | LLM segments min f_min and max f_max | G2 contiguity, model applicability |
| H8 | low-level amplitude 0.5 g and level ratio 0.5 g to 8 g implied by the trained model | LLM characteristic_sweep.amplitude and top acceptance level | G1, G2 low level below acceptance level, G5 control median of characteristic runs, model applicability (2 percent) |
| H9 | excitation axes X Y Z and loop order | LLM acceptance_sine.axes (row order) and wings.*.sensor_directions | G2 subset of XYZ, G5 folders and DOF directions |
| H10 | axis order per wing (+Y XYZ / -Y XZY) | LLM wings.*.axis_order | G2 permutation of acceptance axes, G3, G4, G5 xlsx run creation timestamps |
| H11 | run list per wing and axis (pre characteristic / acceptance / post characteristic, -Y Y has three acceptance runs) | LLM wings.*.axes.*.runs (number of acceptance runs) mapped to files found on disk | G2 starts with pre and ends with post, G5 number of acceptance files equals extracted count |
| H12 | which acceptance run is complete (last run) and which are aborted (-Y Y runs 1 and 2) | LLM wings.*.axes.*.aborted_runs (complete = not aborted) | G2 indices in range and at least one complete run, G3, G4, G5 aborted run ends below band end in xlsx X axis |
| H13 | anomalous channel excluded (-Y point 008 step at the 10-11 Hz level change) | data-derived detector (level-transition jump of complete in-axis acceptance channels, not stated in text) | marked as data-derived |
| H14 | control channel set and averaging (4 channels, average) | LLM control.n_control_points and control.strategy | G1 strategy vocabulary, G5 xlsx Number of control channels and Control strategy |
| H15 | response channels (group Measure) and point ids | xlsx DOF ids (authoritative), LLM wings.*.points used as cross-check | G5 LLM response ids equal xlsx ids |

| Id | Condition | Source in the pipeline | Gates |
|---|---|---|---|
| H16 | wing identity (+Y = first block, -Y) | LLM wings keys mapped to data folders | G2 both wings present, G5 folder exists |
| H17 | panel of each point (outer / mid / inner / yoke) | LLM point name_cn keywords Chinese panel keywords | G5 names match xlsx DOF names |
| H18 | limited (notched) points per axis | data-derived from xlsx Mark curves (not stated in text, LLM value ignored) | G2 must be response point ids (catches control ids) |
| H19 | relevance threshold 20 g for calibration scores | method parameter (not a test condition) | none |
| H20 | mode amplitude threshold 10 g and mode band 8-120 Hz in find_modes | method parameters, A8 uses the config profile | none |

**Gate rules.** Table 10 summarizes the checks of each gate with their tolerances.

**Table 10.** Checks and tolerances of the validation gates.

| Gate | Name | Checks | Tolerance |
|---|---|---|---|
| G1 | Parse and units | value parseable, explicit unit, m/s² to g and peak to peak to zero to peak, converted value rounded to the precision of the source, no default values | exact |
| G2 | Physics | segments contiguous and continuous in level, no transition at either end, positive levels, acceptance band inside the low-level band, low level below acceptance levels, limit over top level in (1, 20], abort limit above limit, axis order a permutation, runs start and end with a low-level sweep, aborted indices valid with one complete run, limited points not control points, control point count | 1% band edges, 5% level |
| G3 | Quotes | value kept by the quote verifier V2 with elided quotes of stage A | verbatim |
| G4 | Cross model | value equal to that of the other model on the same input | 1% numeric |
| G5 | Measurement exports | axes with acceptance files, low-level band and control level, measured control spectrum against the profile in each segment, limit and abort marks, control channel count and strategy, number of acceptance files, runs ending early, axis order from time stamps, point identifiers | 2% band and marks, 5% levels |
| G6 | Applicability | low-level band, low level, acceptance band and top level equal to those of the training data of the bound model | 2% |

**Injected errors.** Table 11 lists all 19 injections with the gates that fired, the outcome under the consensus policy and the effect on WarpNet-CQR at c = 0.95 if the run had not been stopped. Frozen effects use the stored predictions because the fingerprint of the dataset differed.

**Table 11.** Injected errors in the evidence output of DeepSeek V4.1 Flash on the curated excerpt. Effects are those of an unguarded run.

| Injection | Field use | Gates fired | Gates without G5 | All gates | Δ over-notch (oct) | Coverage | Δ under-notch (oct) |
|---|---|---|---|---|---|---|---|
| limit unit omitted | consumed | G1 | STOP | STOP | not run | not run | not run |
| limit 65 to 56 g | consumed | G3, G4, G5 | STOP | STOP | +14.1 | 1.00 | 0.0 |
| limit 65 to 650 g (decimal slip) | consumed | G2, G3, G4, G5 | STOP | STOP | −82.4 | 1.00 | +1.3 |
| limit 637.7 read as g (m/s² slip) | consumed | G2, G3, G4, G5 | STOP | STOP | −82.3 | 1.00 | +1.3 |
| top level 8 to 7 g | consumed | G3, G4, G5, G6 | STOP | STOP | −3.2 | 1.00 | 0.0 |
| 8 to 10 Hz level 2 to 2.5 g | consumed | G2, G3, G4, G5 | STOP | STOP | 0.0 | 1.00 | 0.0 |
| displacement 0-peak read as peak-peak | consumed | G2, G3, G4, G5 | STOP | STOP | 0.0 | 1.00 | 0.0 |
| displacement 7.75 to 15.5 mm 0-peak | consumed | G2, G3, G4, G5 | STOP | STOP | 0.0 | 1.00 | 0.0 |
| all levels m/s² read as g | consumed | G2, G3, G4, G5, G6 | STOP | STOP | +589.3 | 1.00 | 0.0 |
| low level 0.5 to 5 g | consumed | G2, G4, G5, G6 | STOP | STOP | 0.0 | 1.00 | 0.0 |
| axis Z omitted | consumed | G4, G5 | STOP | STOP | −33.8 | 0.58 | +0.9 |
| control average to maximum | consumed | G3, G4, G5 | STOP | STOP | +6.4 | 1.00 | 0.0 |
| control points 4 to 3 | consumed | G3, G4, G5 | STOP | STOP | 0.0 | 1.00 | 0.0 |
| -Y Y aborted runs 1,2 to none | consumed | G4, G5 | STOP | STOP | +0.2 | 1.00 | 0.0 |
| -Y Y aborted runs 1,2 to 1 | consumed | G4, G5 | STOP | STOP | +0.5 | 1.00 | 0.0 |
| -Y Y repeats missed (one run) | consumed | G4, G5 | STOP | STOP | +4.4 | 1.00 | 0.0 |
| -Y axis order XZY to XYZ | audit | G3, G4, G5 | FLAG | FLAG | 0.0 | 1.00 | 0.0 |
| -Y Y limited points set to C1 to C4 | audit | G2, G3, G4 | FLAG | FLAG | 0.0 | 1.00 | 0.0 |
| abort limit 65 g added | audit | G2, G3, G5 | FLAG | FLAG | 0.0 | 1.00 | 0.0 |

**Repairs and policies.** Table 12 counts the stops of the strict and the consensus policy, both with G5, on the E1 outputs at each repair stage, separately for the development inputs, on which the repairs were designed, and the held-out inputs. Stage 0 uses the V2 quote verifier, stage A adds elided quotes, stage A+B adds the targeted re-ask, and the last stage also repairs the output of the other model.

**Table 12.** False stops of the strict policy among correct E1 configurations and stops of the consensus policy among configurations with a wrong consumed value, per repair stage.

| Repair stage | Dev, correct stopped by strict | Dev, wrong stopped by consensus | Held-out, correct stopped by strict | Held-out, wrong stopped by consensus |
|---|---|---|---|---|
| 0, V2 verifier | 1 of 4 | 0 of 0 | 4 of 5 | 1 of 1 |
| A, elided quotes | 0 of 4 | 0 of 0 | 4 of 5 | 1 of 1 |
| A+B, re-ask | 0 of 4 | 0 of 0 | 2 of 6 | 0 of 0 |
| A+B, both models repaired | 0 of 4 | 0 of 0 | 1 of 6 | 0 of 0 |

**Re-ask prompt.** Stage B sends the extraction system prompt with the evidence rules and this user message, followed by the restricted template and the chunk.

```
The quotes previously given for the fields in the template below could not be found verbatim in the report
text, or did not contain the value they support. Re-extract only these fields. Return one JSON object with
exactly the nesting of the template. Every quote must be copied character for character from the report text,
without ellipses and without joining text that is not adjacent. When the label, the range and the value of a
fact sit in different rows or cells, give q as a list of those exact texts so that every stated number is
inside a quote.
```

**Version log.** Table 13 lists the pipeline versions in the order of their development, all on 14 November 2026. Effects are given as the fractions of the configurations that were identical to the gold, escalated and silently wrong.

**Table 13.** Development log of the pipeline.

| Version | Change | Design data | Effect |
|---|---|---|---|
| v0 | no gates, parsed LLM config drives the numerics directly, converted units not rounded | - | 0.80 / 0.00 / 0.05 (Qwen E1 unit shift runs with m/s² values read as g) |
| v1 | gates G1 to G6 with consensus policy and xlsx adjudication, V2 quote verifier of step7_llm | curated and all settings (first build) | 0.80 / 0.40 / 0.00 (strict policy 0.60 / 0.75 / 0.00 with 6 E1 stops) |
| v1.1 | gold.json removed from the pipeline, G6 and the stale-model check read the training artifacts (modelcard.py dataset.npz bands and profile and simulator levels), stale check now compares the whole profile | - | no change in outcomes, stale flag also set when only the 5 to 8 Hz displacement differs |
| v2 | after unit conversion keep the shortest canonical decimal that converts back to the source number at its stated precision (637.7 m/s² to 65 g, 78.5 to 8.0 g, 19.6 to 2.0 g, 4.9 to 0.5 g) | unit shift outputs, rule added after inspecting them | 0.95 / 0.40 / 0.00, the four unit shift configs become bitwise exact except the stopped Qwen E1 one |
| v3 | stage A elided-quote verifier (fragments verbatim and in order), stage B one targeted re-ask of required fields still failing G3 using the chunk they came from, both models repaired before the cross model check | designed on dev (curated and chunked) then frozen, evaluated on held-out (complete record and layout shift and unit shift) | consensus 1.00 / 0.40 / 0.00 (all escalations are flags on audit-only fields), strict 0.95 / 0.60 / 0.00 with 1 E1 stop |
| v3 | lead model retrained on the dataset built from the DeepSeek curated LLM config (same seeds) to test training-stage dependence | - | not bitwise (cuDNN nondeterminism on identical inputs), median prediction difference 5e-5 dex and max 0.013 dex, over-notch at 95 percent 89.04 vs 88.93 octaves, coverage 1.00 and under-notch 0 in both |
| v3 | summary checker v2 assigns every number to wing and axis and quantity from the text and compares with that table cell, prompt v2 asks for one clause per wing and axis with unit words | - | prompt v2 36/36 numbers verified for both models, 30/30 single misattribution mutations detected (membership checker 0/30), free-style prompt v1 texts 23/33 and 11/27 attributable |

# Appendix D. Numerical Method Details

**Simulator.** Synthetic pairs are generated from a modal expansion of the transmissibility,

$$H(f) = c_0 + \sum_{k=1}^{K} \frac{r_k \rho_k^2}{1 - \rho_k^2 + 2i\tilde{\zeta}_k \rho_k}, \rho_k = \frac{f}{\tilde{f}_k} \tag{7}$$

$$\tilde{f}_k = f_k(1 - \delta s_k), \tilde{\zeta}_k = \zeta_k(1 + \eta s_k), s_k = 1 - \exp\left(-\frac{a_k}{a_0}\right), a_k = \frac{G|r_k|}{2\zeta_k} \tag{8}$$

with K from 2 to 8, $f_k$ log-uniform on 12 to 260 Hz, $\zeta_k$ log-uniform on 0.005 to 0.05, $|r_k|$ log-uniform on 0.03 to 1.5 with random signs, $c_0 = 1$ for in-axis channels drawn with probability 0.4 and log-uniform on 0.003 to 0.5 otherwise, with residues of cross-axis channels scaled by a common factor log-uniform on 0.05 to 1. The nonlinearity parameters are $\delta$ uniform on [0, 0.15], $\eta$ uniform on [0, 3] and $a_0$ log-uniform on [0.5, 30] g. Each sample is evaluated at G = 0.5 g and 8 g, converted to $\log_{10} |H|$ and perturbed by noise of 0.01 dex. Training a surrogate on simulated structural responses follows neural operator work in structural engineering [23], and here the network is then fine-tuned on measurements. The two levels are fixed in the simulator, which is why gate G6 rejects campaigns at other levels.

**Virtual units.** For each excitation axis of the training wing the law

$$\Delta \ln f_p =- \alpha s, \Delta \ln T_p =- \ln(1 + \beta s), s = 1 - \exp\left(-\frac{A_8}{a_0}\right) \tag{9}$$

is fitted by least squares to matched resonance peaks, where $A_8$ is the linearly extrapolated peak response. The data show no trend of the shift with $A_8$ (Spearman $\rho = -0.08$, $p = 0.22$), so the fitted law should be read as a parameterization of the mean and spread of the matched shifts rather than as an amplitude dependence. The matching for this fit searched the acceptance-level curve between 0.85 and 1.05 times the low-level frequency, an asymmetric window that biases the matched shifts downward, so the prior is deliberately conservative. A log-normal prior for ($\alpha$, $\beta$, $a_0$) is formed from the three axis-wise fits, with the standard deviation of each log parameter bounded below by 0.3. A virtual unit draws one parameter set per axis and applies it to all low-level curves of the training wing by a piecewise linear warp of the peaks and a Gaussian bump in log amplitude.

**WarpNet head.** The backbone outputs fields $h_1$ to $h_4$, and

$$d(f) = \lambda(f)[-0.03 + 0.18\varphi(h_1(f) - \ln 5)] \tag{10}$$

$$g(f) = \lambda(f)[0.2\tanh h_2(f) - \mathrm{softplus}(h_3(f) - 4)] \tag{11}$$

$$\lambda(f) = \varphi\left(\frac{\ln f - \ln 8}{0.05}\right), \hat{y}^+(f) = \hat{y}(f) + \mathrm{softplus}(h_4(f) - 2) \tag{12}$$

where $\varphi$ is the logistic function and $\bar{d}$ is d smoothed by a 9-point moving average. The warp lies in [−0.03, 0.15] in natural log frequency, the factor $\lambda$ switches both fields off below about 8 Hz where the stowed wing moves rigidly, and the zero-initialized head starts close to the low-level curve.

**Loss and training.** With mask $m_j$ and weight $w_j = 1 + 2 \max(y_j, 0)$ the loss is

$$\mathcal{L} = \frac{\sum_j w_j m_j \left[\rho_{0.5}(y_j - \hat{y}_j) + \rho_{0.9}(y_j - \hat{y}_j^+)\right]}{\sum_j w_j m_j}, \rho_\tau(e) = \max(\tau e, (\tau - 1)e) \tag{13}$$

Three virtual units per training fold are added with a sample weight of 1/3, and the low-level input is smoothed with a Savitzky-Golay filter. Five members are pretrained for 20 epochs and fine-tuned for 150 epochs with AdamW and a cosine schedule.

**Calibration.** Out-of-fold scores come from four folds grouped by response location. With N real and $N_v$ virtual scores weighted 0.5/N and 0.5/$N_v$,

$$\hat{q} = \inf\left\{ q | \sum_n \omega_n \mathbf{1}[s_n \le q] \ge c\left(1 + \frac{1}{N}\right)\right\} \tag{14}$$

For models without a quantile output $\bar{u}$ is the ensemble mean and $\sigma$ the ensemble spread or, for the heteroscedastic head, the predicted standard deviation combined with the ensemble spread. The fixed margin of the reference is x(f) + m/20 with m from 0 to 40 dB.

**Compared variants.** Non-learned baselines are linear extrapolation, a physics mapping M1 with one frequency scale and gain per axis, and a modal-parametric model with gradient-boosted quantile regressors. Neural variants use a FiLM-CNN, a U-Net [37], a patch Transformer [38] or a Fourier neural operator [39] with a residual, heteroscedastic, CQR or WarpNet head, and data variants add pair-consistent augmentation, mixup [40], virtual-unit augmentation, input smoothing or triaxial input.

Figure 8 shows simulated pairs. Over all pairs the largest peak between 8 and 100 Hz moves by a median of −1.0 percent (quartiles −3.8 and +0.0, grid spacing about 1 percent) and its height changes by a median factor of 0.90 (quartiles 0.71 and 0.99). Figure 9 compares the coverage and the over-notch of bounds calibrated with and without virtual-unit scores. Out-of-fold scores alone reached the nominal coverage in 12 of 46 cases at c = 0.90 and 19 of 46 at c = 0.95 (median coverage 0.71 and 0.94), and adding virtual units raised this to 45 and 45 of 46, at a median of 4.9 and 5.8 times the over-notch of calibration without virtual units.

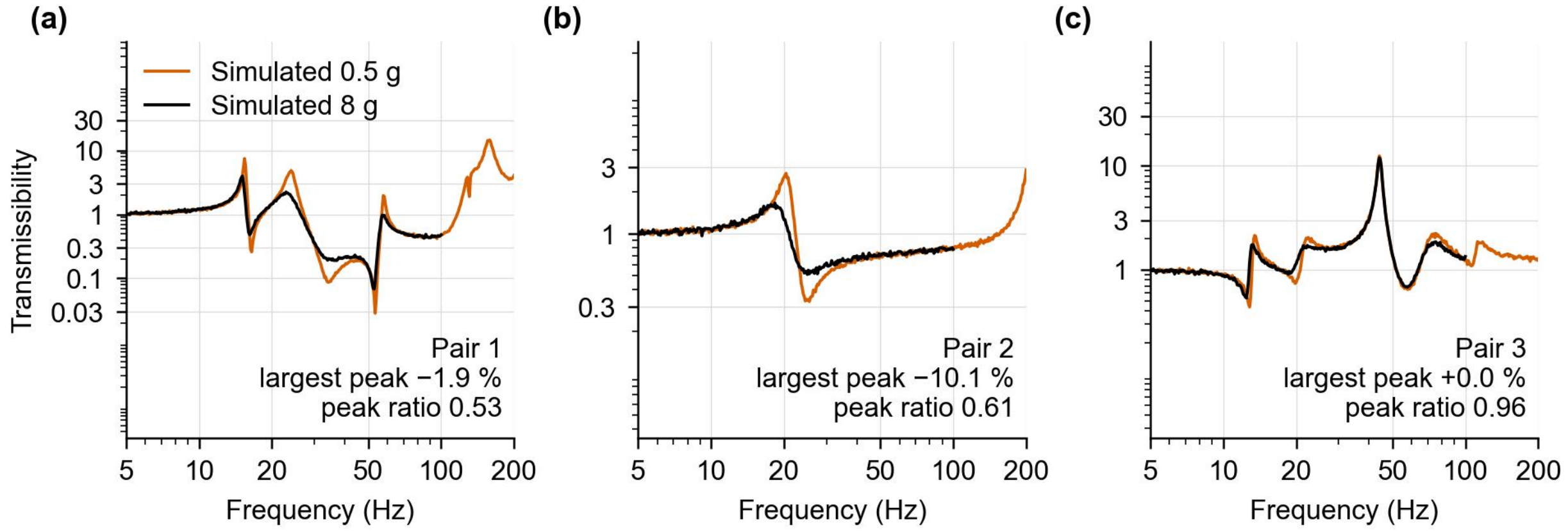

**Figure 8.** Physics based synthetic pairs used for pretraining. (a to c) The first three in-axis pairs of the 20000 simulated pairs, softening and added damping grow with the modal amplitude, so the 8 g resonances move down and flatten.

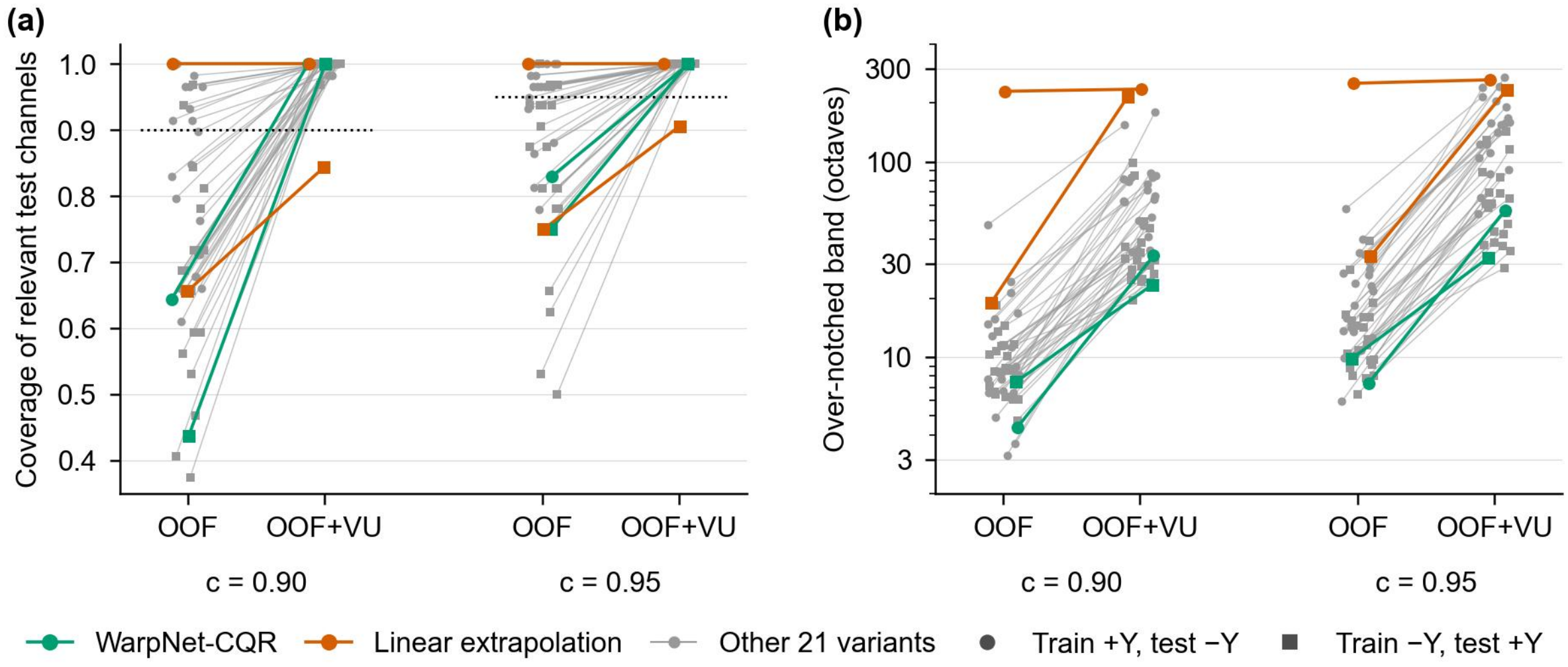

**Figure 9.** Calibration source. Each line joins one of 23 methods in one direction calibrated on the training wing with out-of-fold scores only (OOF) and with out-of-fold and virtual-unit scores (OOF+VU). (a) Achieved coverage of relevant test channels, the dotted lines mark the nominal level. (b) The price in over-notched bandwidth. The physics mapping M1 is not shown.

# Appendix E. Full Numerical Benchmark

**Table 14.** Reduction in percent of the over-notched bandwidth relative to the oracle fixed margin at equal coverage of relevant channels, with 95% bootstrap intervals over response locations. M1 is omitted because it performed far worse than linear extrapolation in both directions.

| Method | logRMSE (dex) | +Y→−Y 90% | +Y→−Y 95% | −Y→+Y 90% | −Y→+Y 95% |
|---|---|---|---|---|---|
| WarpNet-CQR w/o VU aug. and smoothing | 0.091 | 71 [32, 80] | 71 [31, 94] | 96 [71, 98] | 94 [90, 97] |
| FiLM-CNN + heteroscedastic head | 0.087 | 68 [37, 85] | 71 [39, 94] | 96 [59, 97] | 95 [75, 97] |
| WarpNet-CQR + triaxial input | 0.084 | 65 [29, 80] | 69 [30, 93] | 95 [57, 97] | 95 [92, 97] |
| FiLM-CNN + CQR head | 0.090 | 64 [24, 80] | 67 [30, 93] | 96 [63, 98] | 96 [92, 98] |
| WarpNet head w/o quantile output + VU aug. | 0.088 | 70 [38, 84] | 72 [45, 95] | 92 [21, 95] | 87 [82, 94] |
| FiLM-CNN + VU augmentation | 0.087 | 72 [26, 85] | 62 [−23, 93] | 93 [28, 97] | 92 [88, 96] |
| FiLM-CNN + CQR + VU aug. + smoothing | 0.085 | 58 [28, 72] | 69 [38, 94] | 95 [56, 97] | 95 [89, 97] |
| Patch Transformer | 0.096 | 64 [32, 79] | 62 [36, 91] | 93 [56, 97] | 92 [88, 96] |
| **WarpNet-CQR (proposed)** | **0.085** | **55 [16, 78]** | **64 [31, 93]** | **94 [46, 97]** | **94 [87, 97]** |
| WarpNet head w/o quantile output | 0.095 | 56 [4, 76] | 57 [−205, 92] | 93 [23, 96] | 92 [88, 96] |
| Fourier neural operator | 0.103 | 53 [18, 68] | 60 [−12, 77] | 92 [26, 95] | 91 [85, 94] |
| WarpNet head w/o quantile output + triaxial input | 0.091 | 52 [2, 75] | 53 [8, 90] | 94 [43, 96] | 92 [85, 96] |
| WarpNet head w/o quantile output + triaxial + VU aug. | 0.090 | 53 [−2, 72] | 50 [−12, 90] | 91 [22, 94] | 90 [85, 94] |
| FiLM-CNN + input smoothing | 0.092 | 38 [−1, 67] | 49 [7, 91] | 95 [64, 97] | 95 [92, 97] |
| FiLM-CNN (base) | 0.094 | 45 [−10, 65] | 44 [−329, 88] | 95 [47, 97] | 93 [88, 97] |
| U-Net 1D | 0.095 | 42 [1, 73] | 41 [−7, 88] | 95 [57, 97] | 93 [84, 97] |
| Modal-parametric (GBM) | 0.100 | 29 [−20, 57] | 51 [4, 89] | 95 [54, 97] | 95 [91, 97] |
| FiLM-CNN w/o pair augmentation | 0.094 | 41 [−20, 62] | 37 [−182, 86] | 95 [58, 97] | 94 [90, 97] |
| FiLM-CNN + mixup | 0.091 | 29 [−36, 74] | 29 [−25, 80] | 95 [58, 97] | 94 [88, 97] |
| FiLM-CNN w/o pretraining | 0.094 | −2 [−52, 45] | 21 [−67, 78] | 91 [13, 95] | 83 [74, 95] |
| FiLM-CNN + triaxial input | 0.095 | 22 [−119, 54] | −29 [−320, 76] | 96 [61, 98] | 95 [91, 97] |
| WarpNet head w/o quantile output, w/o pretraining | 0.102 | −57 [−222, 16] | −29 [−286, 75] | 82 [−10, 91] | 66 [57, 88] |
| Linear extrapolation | 0.101 | −2 [−3, 2] | −3 [−7, 2] | 0 [−18, 4] | 0 [−1, 0] |

**Table 15.** Total over-notched bandwidth in octaves of bounds calibrated on the training wing for all variants except M1, with the achieved coverage in parentheses, sorted by the sum over both directions at c = 0.90.

| Method | +Y→−Y c = 0.90 | +Y→−Y c = 0.95 | −Y→+Y c = 0.90 | −Y→+Y c = 0.95 |
|---|---|---|---|---|
| Modal-parametric (GBM) | 35.4 (1.00) | 65.5 (1.00) | 19.6 (1.00) | 28.6 (1.00) |
| **WarpNet-CQR (proposed)** | **33.1 (1.00)** | **56.7 (1.00)** | **23.5 (1.00)** | **32.3 (1.00)** |
| FiLM-CNN + CQR + VU aug. + smoothing | 35.3 (1.00) | 60.8 (1.00) | 24.8 (1.00) | 35.0 (1.00) |
| WarpNet-CQR + triaxial input | 34.7 (1.00) | 54.2 (1.00) | 25.8 (1.00) | 37.4 (1.00) |
| WarpNet-CQR w/o VU aug. and smoothing | 47.4 (1.00) | 91.0 (1.00) | 24.3 (1.00) | 37.0 (1.00) |
| FiLM-CNN + CQR head | 52.0 (1.00) | 110.8 (1.00) | 26.8 (1.00) | 38.4 (1.00) |
| FiLM-CNN + VU augmentation | 49.9 (1.00) | 105.0 (1.00) | 29.1 (1.00) | 43.8 (1.00) |
| Patch Transformer | 64.4 (1.00) | 243.5 (1.00) | 25.0 (1.00) | 69.3 (1.00) |
| FiLM-CNN + mixup | 66.9 (1.00) | 142.1 (1.00) | 24.1 (1.00) | 42.5 (1.00) |
| WarpNet head w/o quantile output + VU aug. | 41.5 (1.00) | 86.5 (1.00) | 49.7 (1.00) | 83.0 (1.00) |
| WarpNet head w/o quantile output | 71.5 (1.00) | 154.3 (1.00) | 29.4 (1.00) | 47.9 (1.00) |
| U-Net 1D | 62.5 (1.00) | 123.5 (1.00) | 38.4 (1.00) | 65.1 (1.00) |
| WarpNet head w/o quantile output + triaxial input | 77.2 (1.00) | 191.9 (1.00) | 31.4 (1.00) | 60.8 (1.00) |
| WarpNet head w/o quantile output + triaxial + VU aug. | 63.3 (1.00) | 123.5 (1.00) | 45.9 (1.00) | 88.1 (1.00) |
| Fourier neural operator | 83.1 (1.00) | 271.7 (1.00) | 28.0 (1.00) | 69.7 (1.00) |
| FiLM-CNN w/o pair augmentation | 79.5 (1.00) | 158.5 (1.00) | 32.0 (1.00) | 57.9 (1.00) |
| FiLM-CNN (base) | 81.3 (0.98) | 159.8 (1.00) | 34.1 (1.00) | 61.1 (1.00) |
| FiLM-CNN + input smoothing | 83.5 (1.00) | 157.9 (1.00) | 33.4 (1.00) | 54.6 (1.00) |
| FiLM-CNN + triaxial input | 87.7 (0.98) | 167.5 (1.00) | 31.2 (1.00) | 55.7 (1.00) |
| FiLM-CNN + heteroscedastic head | 84.8 (1.00) | 205.6 (1.00) | 36.8 (0.97) | 115.7 (1.00) |
| WarpNet head w/o quantile output, w/o pretraining | 154.2 (1.00) | 215.2 (1.00) | 98.9 (1.00) | 143.1 (1.00) |
| FiLM-CNN w/o pretraining | 180.7 (1.00) | 240.7 (1.00) | 84.8 (1.00) | 129.8 (1.00) |

| Method | +Y→−Y c = 0.90 | +Y→−Y c = 0.95 | −Y→+Y c = 0.90 | −Y→+Y c = 0.95 |
|---|---|---|---|---|
| Linear extrapolation | 235.8 (1.00) | 263.0 (1.00) | 214.7 (0.84) | 234.4 (0.91) |

Tables 14 and 15 complete the comparison of Section 7. WarpNet-CQR has the lowest calibrated cost of all variants in the +Y→−Y direction at c = 0.90, whereas in the −Y→+Y direction the modal-parametric baseline is lower (19.6 and 28.6 octaves). Along the oracle frontier in the +Y→−Y direction WarpNet-CQR reduces the over-notch of a fixed margin by 55% and 64% at 90% and 95% coverage and FiLM-CNN with a CQR head by 64% and 67%. Simulator pretraining is essential. Without it, the calibrated FiLM-CNN over-notches 181 and 85 octaves at c = 0.90 instead of 81 and 34 octaves, and in the +Y→−Y direction the WarpNet head without quantile output falls from 56% and 57% to −57% and −29%. A quantile output is the most useful single change, reducing the calibrated over-notch of FiLM-CNN from 81 and 34 octaves to 52 and 27 octaves. The heteroscedastic head improves the frontier (68% and 71%) but not the calibrated bound. Virtual-unit augmentation gives the largest gain for calibrated bounds, from 81 and 34 octaves to 50 and 29 octaves for FiLM-CNN, while smoothing alone (84 and 33 octaves) and pair-consistent augmentation (79 and 32 octaves without it) change little. Mixup narrows the calibrated bound of FiLM-CNN (67 and 24 octaves) but lowers its frontier reduction to 29% and 29%. Triaxial input does not consistently improve the calibrated bounds of either model. The patch Transformer (64% and 62%) and the Fourier neural operator (53% and 60%) improve on the base FiLM-CNN (45% and 44%) along the frontier, the U-Net does not (42% and 41%), and none reaches the quantile models. The two comparisons rank methods differently because virtual-unit augmentation narrows the calibrated bound but gives a worse trade-off along the oracle frontier. The global physics mapping M1 performed far worse than linear extrapolation in both directions.

**Table 16.** Achieved coverage of relevant test channels for bounds calibrated with out-of-fold scores only (OOF) or with additional virtual-unit scores (OOF+VU), for +Y→−Y / −Y→+Y.

| Method | OOF c = 0.90 | OOF c = 0.95 | OOF+VU c = 0.90 | OOF+VU c = 0.95 |
|---|---|---|---|---|
| **WarpNet-CQR (proposed)** | **0.64 / 0.44** | **0.83 / 0.75** | **1.00 / 1.00** | **1.00 / 1.00** |
| WarpNet-CQR w/o VU aug. and smoothing | 0.61 / 0.69 | 0.98 / 0.81 | 1.00 / 1.00 | 1.00 / 1.00 |
| FiLM-CNN + CQR head | 0.66 / 0.53 | 1.00 / 0.75 | 1.00 / 1.00 | 1.00 / 1.00 |
| FiLM-CNN + heteroscedastic head | 0.92 / 0.66 | 1.00 / 0.78 | 1.00 / 0.97 | 1.00 / 1.00 |
| Modal-parametric (GBM) | 0.68 / 0.78 | 0.78 / 0.94 | 1.00 / 1.00 | 1.00 / 1.00 |
| FiLM-CNN (base) | 0.92 / 0.84 | 0.97 / 0.94 | 0.98 / 1.00 | 1.00 / 1.00 |
| Linear extrapolation | 1.00 / 0.66 | 1.00 / 0.75 | 1.00 / 0.84 | 1.00 / 0.91 |

Table 16 shows that out-of-fold calibration alone under-covers in at least one direction for most learned models, because held-out channels of the training wing share its unit-level deviation. With virtual-unit scores all learned models reach a coverage not lower than the nominal level in both directions, close to 1. With 59 and 32 relevant test channels, coverage values move in steps of about 0.02 and 0.03.

The error of linear extrapolation grows with the local slope of the low-level curve, with a positive per-channel Spearman correlation in 124 of 126 in-axis channels (median $\rho = 0.28$, sign test $p = 2 \times 10^{-34}$) and a binned median error rising monotonically from −0.11 to +0.08 dex. Rising flanks are under-predicted and falling flanks over-predicted, as expected when resonances shift to lower frequencies. The largest single under-predictions occur at antiresonance troughs. An amplitude margin cannot represent such a frequency-dependent error, whereas a warp of the frequency axis can.

Figure 10 summarizes the ablation of the bound model at both calibration levels. Removing pretraining on simulated pairs multiplies the over-notch by 1.4 to 3.4 in all 8 comparisons of both families, directions and levels. Adding the CQR head changes it by a factor of 0.59 to 0.83, adding virtual unit augmentation and input smoothing by 0.55 to 0.97, and a triaxial input by 0.91 to 1.16, which lowers the over-notch in only 3 of the 8 comparisons.

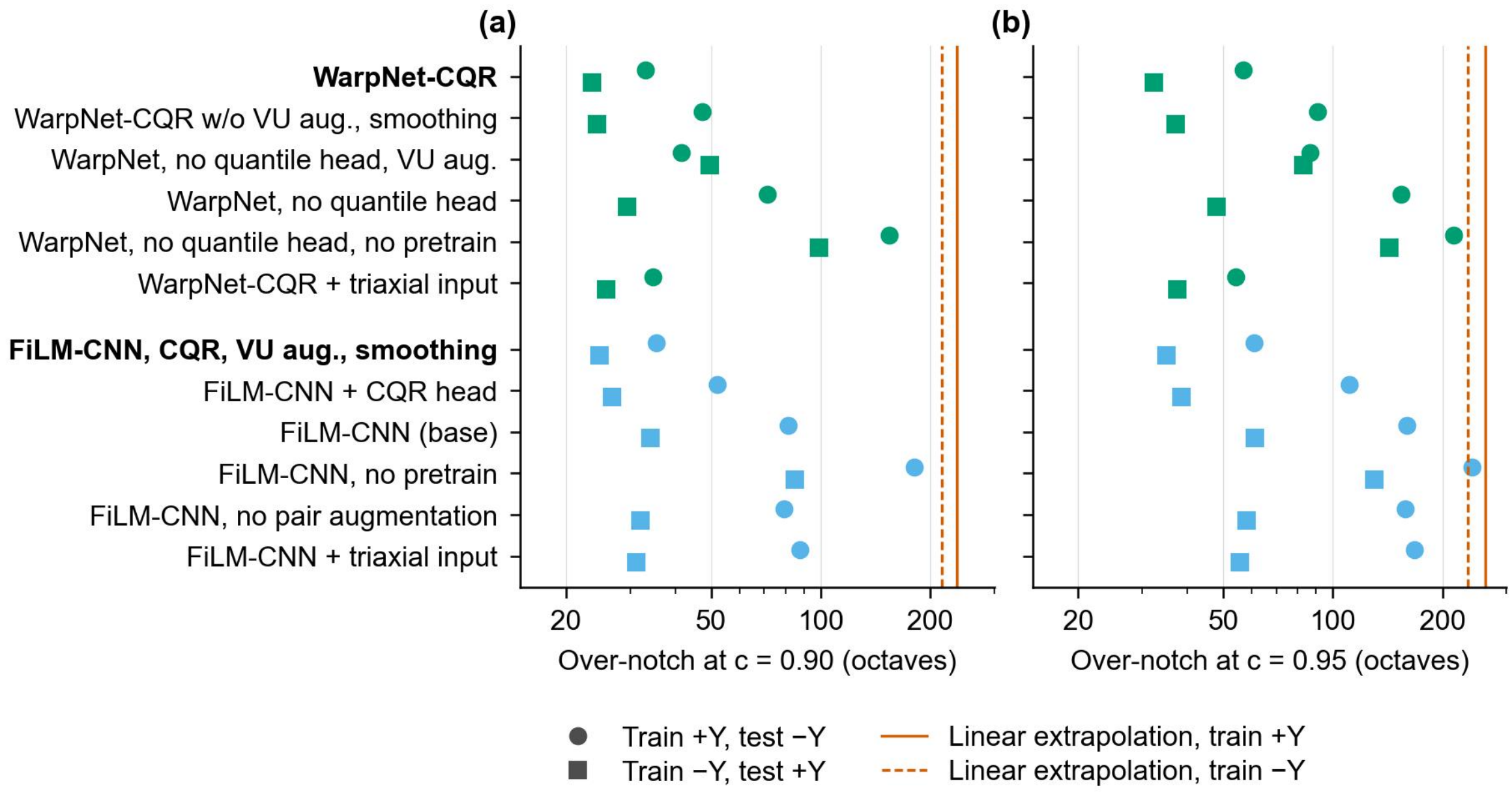

**Figure 10.** Ablation of the bound model, total over-notched bandwidth on the test wing of bounds calibrated on the training wing at c = 0.90 (a) and c = 0.95 (b), WarpNet variants in green and FiLM-CNN variants in blue, circles train +Y and test −Y, squares the reverse. Vertical lines mark linear extrapolation calibrated the same way. The achieved coverage of relevant test channels reaches the nominal level in 48 of 48 cases.

# Appendix F. Significance of the In Context Comparison

To test whether the differences of Section 8 exceed the variation between channels, we compared methods in pairs on the same channels, with a paired bootstrap of the pooled logRMSE difference over channels with 10000 resamples and with a Wilcoxon signed rank test on the per-channel logRMSE, and corrected the p-values of the 12 direction-wise comparisons with the Holm procedure. Only the gaps between the LLMs without examples and WarpNet-CQR remained significant at the 5% level, in both directions and under both tests, with differences of 0.021 to 0.039 dex. Over both directions DeepSeek V4.1 Flash without examples differed from linear extrapolation on 48 points by 0.0009 dex (95% interval 0.0002 to 0.0018), so the two are practically equivalent. Qwen-plus without examples was 0.0057 dex worse than linear extrapolation, with an uncorrected bootstrap p of 0.013 in the −Y→+Y direction that rose to 0.10 after correction. Neither the change of DeepSeek V4.1 Flash from k = 0 to k = 4 nor the gap between the two LLMs at k = 4 was significant after correction. Channels of one wing share the structure and the test run and are not independent, so the intervals and p-values are optimistic.

**Table 17.** Paired comparisons of the in context probe with 10000 bootstrap resamples over channels and a Wilcoxon signed rank test per channel, with Holm correction over the 12 direction-wise comparisons for each test. A bootstrap p below the resolution is entered at its floor for the correction.

| Comparison A vs B | Direction | Channels | logRMSE A minus B | 95% interval | p bootstrap | Holm | p Wilcoxon | Holm |
|---|---|---|---|---|---|---|---|---|
| DeepSeek V4.1 Flash k = 0 vs linear, 48 points | +Y→−Y | 30 | +0.0015 | +0.0003 to +0.0030 | 0.028 | 0.199 | 0.058 | 0.404 |
| DeepSeek V4.1 Flash k = 0 vs linear, 48 points | −Y→+Y | 30 | +0.0006 | −0.0002 to +0.0016 | 0.199 | 0.795 | 0.360 | 1.000 |
| DeepSeek V4.1 Flash k = 0 vs k = 4 | +Y→−Y | 30 | −0.0052 | −0.0195 to +0.0037 | 0.432 | 0.865 | 0.903 | 1.000 |
| DeepSeek V4.1 Flash k = 0 vs k = 4 | −Y→+Y | 30 | −0.0290 | −0.0987 to +0.0095 | 0.574 | 0.865 | 0.137 | 0.687 |
| Qwen-plus k = 4 vs DeepSeek V4.1 Flash k = 4 | +Y→−Y | 26 | +0.1609 | −0.0022 to +0.3282 | 0.043 | 0.256 | 0.040 | 0.317 |
| Qwen-plus k = 4 vs DeepSeek V4.1 Flash k = 4 | −Y→+Y | 29 | +0.0980 | −0.0563 to +0.2557 | 0.208 | 0.795 | 0.620 | 1.000 |
| DeepSeek V4.1 Flash k = 0 vs WarpNet-CQR | +Y→−Y | 30 | +0.0220 | +0.0142 to +0.0307 | $< 1 \times 10^{-4}$ | 0.001 | $< 1 \times 10^{-4}$ | 0.000 |

| Comparison A vs B | Direction | Channels | logRMSE A minus B | 95% interval | p bootstrap | Holm | p Wilcoxon | Holm |
|---|---|---|---|---|---|---|---|---|
| DeepSeek V4.1 Flash k = 0 vs WarpNet-CQR | −Y→+Y | 30 | +0.0310 | +0.0219 to +0.0422 | $< 1 \times 10^{-4}$ | 0.001 | $< 1 \times 10^{-4}$ | 0.000 |
| Qwen-plus k = 0 vs WarpNet-CQR | +Y→−Y | 26 | +0.0213 | +0.0116 to +0.0318 | $< 1 \times 10^{-4}$ | 0.001 | 0.001 | 0.012 |
| Qwen-plus k = 0 vs WarpNet-CQR | −Y→+Y | 29 | +0.0385 | +0.0275 to +0.0525 | $< 1 \times 10^{-4}$ | 0.001 | $< 1 \times 10^{-4}$ | 0.000 |
| Qwen-plus k = 0 vs linear, 48 points | +Y→−Y | 26 | +0.0036 | +0.0001 to +0.0082 | 0.073 | 0.363 | 0.980 | 1.000 |
| Qwen-plus k = 0 vs linear, 48 points | −Y→+Y | 29 | +0.0074 | +0.0026 to +0.0138 | 0.013 | 0.101 | 0.084 | 0.502 |

Positive differences mean that method A has the larger error.

## Appendix G. Key Counts and How They Relate

### Measurement points and channels

- Sensor numbering follows the LMS test system, 1 to 23 with number 15 unused
- The +Y wing table lists 22 response points, the −Y wing 21 (no point 20), plus 4 control channels C1 to C4 per wing
- Each response point is a triaxial accelerometer and contributes 3 response channels
- Exports cover all 22 points of the +Y wing and 20 of the −Y wing, point 008 of the −Y wing is excluded as the anomalous channel of Section 5

### Curve pairs

- A pair joins the low-level and acceptance-level curves of one response channel under one excitation axis for one acceptance run
- +Y wing, 22 points × 3 directions × 3 axes × 1 run = 198 pairs, all from complete runs
- −Y wing, 60 pairs per axis and run, 60 for X, 60 for Z and 180 for Y, the Y axis having been run three times, twice aborted and once completing the upper band, total 300
- 498 pairs in total, 378 from complete runs, the three Y-axis runs of the −Y wing together give one complete coverage of its 60 channels

### Test channels of the bound model

- The test channels of one cross-wing direction are the complete pairs of the test wing, 180 for +Y→−Y and 198 for −Y→+Y
- A channel is relevant when its true acceptance-level response exceeds the 20 g threshold in at least one evaluated bin, 59 of the 180 and 32 of the 198

### Schema and extraction scores

- 76 fields in 10 groups, 32 numeric, 19 categorical and 25 lists
- The gold configuration assigns non-null values to 64 fields, each carrying one of the 64 evidence spans of Section 4.1
- 224 atomic facts, a list contributes one fact per element
- The agreement script counts 236 facts because it also counts each of the 12 null fields
- The gold passed 51 automated checks, and Table 1 adds 9 consistency checks against the measurement files

### Configurations and errors

- 20 configurations, 5 inputs × 2 modes × 2 models
- Table 5 adds the 4 stored outputs before the re-ask, 24 outputs of which 23 are correct
- 1 real error in a consumed field (the unit slip of Section 6.5) and 19 injected errors, 16 in consumed fields and 3 in audit fields
- The gates stop all 17 errors in consumed fields, 16 of which would have run without gates
- The 8 escalations of the final policy are 12 audit-field flags of Qwen-plus on 8 configurations, its abort threshold on 7, the −Y axis order on 3 and the limited-points list on 2

### Pipeline structure

- 6 gates per field, 3 corroboration policies, 2 repairs before the gates decide and 2 quote verifier levels
- Versions v0 to v3 are the pipeline without gates, with gates, with the source precision rule and with the repairs

- Of the 20 hard-coded conditions of the earlier analysis code, 14 are read from the records (2 for audit only), 3 come from the measurement exports, 2 are method parameters and 1 is an implementation detail

**Bound model**

- 5-member ensemble, 178,308 parameters per member
- Pretrained on 20000 simulated pairs, calibrated from the out-of-fold scores of the training wing and of 30 virtual units
- 8 calibrated settings, 2 directions × 2 coverage levels × 2 score sets
- 204 matched modes in the mode statistics
- The error near the dominant mode is compared on the 165 test channels with at least 3 evaluated bins in that window
- The example channel of Figure 5 is chosen among the 20 in-axis channels whose dominant mode would exceed the 65 g limit at 8 g

**Forecasting probe**

- Each curve is resampled to 48 log-spaced points
- The first 30 relevant channels of each direction are tested, 120 requests per model for 2 directions × 2 values of k
- DeepSeek V4.1 Flash answered all 120 requests and Qwen-plus 115, methods are compared on the 26 and 29 channels answered by both
- The paired tests of Appendix F use 10000 bootstrap resamples and correct 12 direction-wise comparisons

**API usage**

- 341 requests and 812,923 tokens for extraction, verification and forecasting
- 10 further requests for the pipeline stage, 351 requests and 843,374 tokens in total, about 0.15 USD

# Appendix H. Terminology

**Table 18.** Terminology used in this paper, ordered by first appearance in the text.

| Term | Definition |
|---|---|
| Notching | Reducing the input level in the frequency bands where a response would exceed its allowable level, and the resulting input profile |
| Low-level sweep | The 0.5 g sine sweep run before and after each acceptance-level sweep to characterize the structure |
| Acceptance level | The full test level of the campaign, reaching 8 g in the 11 to 100 Hz band |
| Response point | One triaxial accelerometer location, contributing three response channels, also called a measurement point |
| Control channel | One of the four accelerometers C1 to C4 whose average defines the applied input |
| Transmissibility | The ratio of a response amplitude to the average amplitude of the four control channels, as a function of frequency |
| Test records | The Chinese textual records of the campaign, the input to the extractors |
| Measurement exports | The curve files exported from the LMS test system |
| Schema | The set of 76 typed fields that an extractor must fill |
| Configuration | An assignment of a value or null to every schema field, written $\hat{c}$ |
| Gold configuration | The reference configuration c*, checked against the records and the export metadata |
| Atomic fact | A non-null scalar field or one element of a non-null list, the unit of the extraction score |
| Consumed field | A field read by the numerical stage, in which an error can change the result |
| Audit field | A field kept only for checking, whose failure raises a flag without stopping the run |
| Gate | A deterministic check placed between the LLM output and the numerical stage, G1 to G6 |
| Escalation | A run that stops or flags at least one field for a human |
| False stop | A stop of a correct configuration |
| Silent error | A run that completes with a wrong value in a consumed field |
| Exact agreement | Downstream results equal to the gold run bit for bit |
| Rounding agreement | Over-notch and under-notch totals within 0.05 octave and coverage within 0.01 of the gold run |
| Input variants | The curated excerpt, its chunked form, the complete records and the layout and wording perturbations |
| Plain mode | Extraction without quotes, the default setting |
| Evidence mode (E1) | Extraction in which every non-null value must carry a verbatim quote |
| Quote verifier (V1, V2) | The deterministic checks of the quotes, generic (V1) plus field-specific keyword rules (V2) |
| Cross model checking (E2) | Each model flags unsupported or wrong fields in the output of the other model |
| Repeat runs (E3) | A second identical call used to measure the stability of the outputs |

| Term | Definition |
|---|---|
| Corroboration policy | The rule that combines the gate outcomes, lenient, consensus (the default) or strict |
| Repair | Stage A extends the verifier to elided quotes and stage B re-asks the same model for the fields whose quotes still fail |
| Pipeline versions | v0 without gates, v1 with gates, v2 with the source precision rule and v3 with the repairs |
| Fingerprint | A hash of everything the trained models saw, used to mark a configuration as stale |
| Relevant channel | A test channel whose true acceptance-level response exceeds the 20 g relevance threshold in at least one evaluated bin, the relevant set collects those bins |
| Coverage | The fraction of relevant test channels whose bound holds on the whole relevant set, at nominal level c |
| Over-notched bandwidth | The cost of a bound, the notched frequency width summed over all test channels in octaves |
| Bound model | The model that maps a low-level curve to an upper bound of the acceptance-level curve, here WarpNet-CQR |
| Conformal calibration | The procedure that sets the bound parameter from held-out scores so that coverage reaches its nominal level |
| Virtual unit | A simulated unit sharing one random nonlinearity, used in training and calibration |
| Cross-wing direction | Training on the pairs of one wing and testing on the complete pairs of the other, written +Y→−Y or −Y→+Y |
| dex | One decade, the unit of errors in $\log_{10}$ transmissibility |
| In-axis channel | A response channel whose sensitive direction coincides with the excitation axis |
| logRMSE | The root mean squared error in $\log_{10}$ transmissibility, in dex |
| k | The number of example curve pairs given in the forecasting prompt, 0 or 4 |
| Oracle frontier | The coverage and over-notch trade-off obtained when the bound parameter is chosen on the test wing itself |